\documentclass[letterpaper, 10 pt, conference]{ieeeconf}  

\IEEEoverridecommandlockouts                              

\usepackage{mathptmx} 
\usepackage{times} 
\usepackage{amsmath} 
\usepackage{amssymb}  
\usepackage{tabularx} 
\usepackage{graphicx}
\usepackage{comment}
\usepackage{mathtools}
\usepackage{upgreek}
\usepackage{gensymb}

\usepackage[noadjust]{cite}
\usepackage{placeins}
\usepackage{xcolor}
\usepackage{url}
\usepackage{mwe}
\usepackage{balance}
\usepackage{multirow}
\usepackage{amsmath}
\usepackage{subcaption}
\usepackage{wrapfig}

\usepackage{dblfloatfix}
\newcolumntype{L}[1]{>{\raggedright\let\newline\\\arraybackslash\hspace{0pt}}m{#1}}
\newcolumntype{C}[1]{>{\centering\let\newline\\\arraybackslash\hspace{0pt}}m{#1}}
\newcolumntype{R}[1]{>{\raggedleft\let\newline\\\arraybackslash\hspace{0pt}}m{#1}}

\newcommand{\altmathcal}[1]{\mathcal{#1}}

\usepackage{mathtools}
\usepackage{wasysym}
\usepackage{bm}
\usepackage{bbm}
\usepackage{booktabs}
\usepackage{float}
\usepackage{cuted}

\usepackage[font=footnotesize,labelfont=bf]{caption}

\usepackage{algpseudocode}
\usepackage[linesnumbered,ruled,vlined,noend]{algorithm2e}
\SetKwRepeat{Do}{do}{while}

\DeclarePairedDelimiter\abs{\lvert}{\rvert}%
\DeclarePairedDelimiter\norm{\lVert}{\rVert}%

\makeatletter
\let\oldabs\abs
\def\abs{\@ifstar{\oldabs}{\oldabs*}}
\let\oldnorm\norm
\def\norm{\@ifstar{\oldnorm}{\oldnorm*}}
\makeatother

\title{\LARGE \bf
Safe Physical Human-Robot Rolling Technique on Realistic Human Models using Reinforcement Learning
}

\author{Diego Williams$^{1\dag}$, Elizabeth Peiros$^{1\dag}$, Michael Owens$^{2}$, \\ Soofyian Atar$^{1}$,  Yuheng Zhi$^1$, Chris D'Ambrosia$^3$ Ananya Rao$^1$, Michael C. Yip$^1$ \IEEEmembership{Senior Member, IEEE}
\thanks{$^\dag$ denotes equal contribution}%
\thanks{$^1$ Elizabeth, Diego, Soofiyan, Yuheng, Ananya and M.C. Yip are with the Electrical and Computer Engineering Department, University of California, San Diego, La Jolla, CA 92093 USA. {\tt\footnotesize\{dec004, epeiros, yip\}@ucsd.edu}}%
\thanks{$^2$ Michael Owens is with the Math and Computer Science Department, University of California, San Diego, La Jolla, CA 92093 USA.}
\thanks{$^3$ Chris D'Ambrosia is with the Mechanical and Aerospace Engineering Department, University of California, San Diego, La Jolla, CA 92093 USA.{\tt\footnotesize \{lyager, edl003, kew054, zsamuels\}@ucsd.edu}}%
}

\begin{document}

\maketitle
\thispagestyle{empty}
\pagestyle{empty}


\begin{strip}
\vspace{-30mm}
    \centering
    \includegraphics[width=\textwidth]{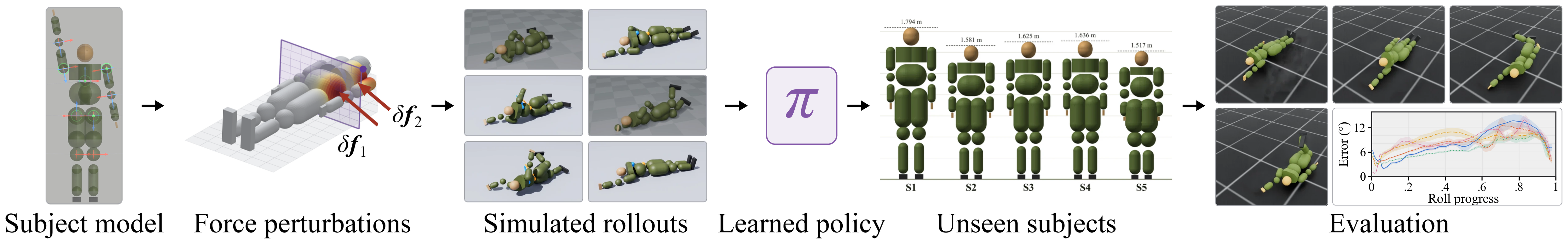}
    \captionof{figure}{Pipeline for safe physical human-robot rolling technique on realistic human models.}
    \label{fig:title-image}
    \vspace{-5mm}
\end{strip}

\begin{abstract}
Rolling and maneuvering patients is a critical task required in healthcare and in field medicine. While robots have demonstrated an ability to roll human-shaped objects, there has not yet been a work to define safe policies for realistic interactions. In this work, we developed a policy for safely rolling patients or injured persons with a realistic human model. The model includes mass, all major body joints, scalability, mass distribution, and inertia properties defined through biomechanics literature. Our policy properly constrains the desired motion of the human during the roll to protect the spinal column, the lumbar-thoracic region, which is the focal point for safety in these types of maneuvers. Our model also considers the safety of any joints known to be injured and penalizes motion of injured joints in the policy development. The developed policies with these safety considerations were deployed on 5 real human models and were able to execute safe successful rolling maneuvers 70.2\% of the time. With injury awareness, the policies were able to reduce the motion of the injured joint by a maximum of 31\%.
\end{abstract}

\section{Introduction}
\vspace{-1mm} 

Although it may seem trivial, rolling and lifting patients can be pivotal to patient care. During the COVID-19 pandemic, hospitalized patients, notably those on ventilation, were rolled prone (face down) for many hours a day to improve oxygenation \cite{paul2020proning}. 
Military corpsmen, in their initial assessment of a wounded soldier, must roll the patient to check for exit wounds or roll the patient from prone to supine (face up) in order to check the airway and other vital signs. 
Unlike other medical tasks, this task carries a risk of musculoskeletal injury to the medical professional. 
Carrying humans to safety was one of the biggest concerns for firefighters' workplace injury~\cite{nazari2020prevalence,aurangabadkar2019prevalence}. In all of these scenarios, medical professionals and first responders must take on personal risk of illness, musculoskeletal injury, or worse in the case of a soldier or a firefighter performing a dangerous rescue operation \cite{trinkoff2003workplace, davis2015prevalence, apple2021ergonomic}. 

While a rescue operation is a series of complex tasks, rolling tasks have clear documentation and strategy for example, the log roll. 
The log roll emphasizes spine stability when transitioning someone from a prone (face down) to a supine (face up) position, or vice versa~\cite{JTS2020CervicalSpine}. 
Robots could use this to deliver time-sensitive patient maneuvering and care when the danger is too great for human medics to assist patients, reducing MSDs (musculoskeletal disorders) and burnout from overworked medical personnel~\cite{galanis2025influence}. 

Today robots still lack the perception capabilities to autonomously assess and interact with complex, potentially fragile patients. Patient dynamics are complex, with multiple uncontrollable articulating limbs, intermittent contact with the surroundings, and high variation in body morphology. 
In more advanced pHRI fields, such as physical therapy and exoskeletons, human-in-the-loop control or compliance via impedance controllers are used to execute tasks, which undermines the development of robotic autonomy to reason and react as opposed to defaulting to being "weak" to remain safe.

Current work on autonomous patient assistance that tackles rolling and lifting for patient transfer tasks requires the patient to self-move to a side-lying posture or bend a knee for the robot to properly execute the task, or use imitation learning and show high-quality real-world tests on unrealistic phantoms \cite{matsumura2025automated, miyake2025deep, Madan2026TACTIC}. 
Although this is encouraging, phantoms with realistic human weight, compliance, and articulation would provide more insight into solutions that translate to real-world task success, which is why we focus on these attributes and accurate spinal motion.

In this paper, we present a strategy and simulated execution of a spine-stable log rolling maneuver that moves a body from prone to supine, or vice versa. Our key contributions include:

\begin{itemize}
    \item Integration of biomechanical safety constraints related to spinal stability, injury considerations, and safe rolling technique as penalties within the reward function
    \item Use of human models with a representatively high degrees of freedom and accurate biomechanical parameters and an inclusion of domain randomization based on anthropomorphic data sets.
    \item Validation of the generalization of the policy on models scaled and modified to real human scans outside the averages ascribed in the domain randomization.
\end{itemize}


\section{Background}
\label{sec: background}

Wilderness medicine protocols emphasize spinal stability during transport and patient repositioning, as destabilizing the spinal column can have significant consequences~\cite{JTS2020CervicalSpine, Blair2012SpineBattle, Cross2011BattlefieldOrtho}. Consequences include the deviation of a compromised spine, leading to paralysis, significantly impacting quality of life, or death. 

These protocols include detailed guides for rolling maneuvers in measured steps. While alternative methods and tools exist, vacuum spine boards and standard NATO litters~\cite{JTS2020CervicalSpine, Rahmatalla2018ThoracicLumbar, Rahmatalla2019Cervical, Swartz2018CervicalMotion, Pernik2016TissuePressure}, the log roll technique is the default patient mobility technique for prone patients and medical teams with fewer than 5 personnel, a common occurrence in combat and wilderness medicine \cite{conrad2012eliminating}. This technique is taught by the National Outdoor Leadership School (NOLS), a premier provider of wilderness medicine training with training across 6 continents.

The guidance to roll a person over solo are as follows: (1) Locate yourself on the left or right side of the patient at approximately their waist (2) move the arm closest to you into a position where the arm is straight and the bicep is against their ear to avoid rolling the person over their shoulder or arm (3) grasp across the person reaching over their back grasping the anterior side of the shoulder girdle and the anterior side of the pelvic bone (iliac crest) (4) rotate the body towards you maintaining rotation of the hips relative to the shoulders thereby minimizing spinal rotation (5) Once the person being rolled reaches their side the hand grip will switch to the posterior side of each of these locations for a slowed descent to the ground. Maintaining spinal rigidity is challenging and requires coordination when rolling patients, as out-of-sync movements can twist, deviate, or compress the spinal column and cause injury.

\section{Related Works}

Few works have begun tackling rolling maneuvers; most use imitation learning~\cite{miyake2025deep}, a simplified human model~\cite{matsumura2025automated}, and/or plush human phantoms made of rigid foam or a plastic shell~\cite{Madan2026TACTIC}. While these works have made meaningful steps, none consider a higher-dimensional human model, particularly one with spinal motion and appropriate weight.

The problem considers the human as motionless and non-responsive, unable to provide assistance during the task, the problem becomes that of learning a control policy for safe manipulation. Some examples of integrating safety constraints in RL for manipulation tasks include~\cite{huang1903learning,lin2019reinforcement}. Some that consider human pose estimation generally address dressing assistance~\cite{wang2023one,ildefonso2021exploiting}


This work considers human modeling. The two most common modeling software include the work from \cite{delp2007opensim}, which created OpenSim, and Anybody Modelling System (AMS) (Aalborg University), Semmie Cad®, Catia®, Jack® (University of Pennsylvania), and Ramsis®. Although these are state-of-the-art biomechanics models that meet the criteria described in \cite{pandy2001computer}, they do not provide outer layers for world contact across the entire body surface and require significant effort to scale the models for user-specificity. In this work, these models would be insufficient for collision dynamics and overly complex in their inclusion of muscle attachments and bones. Thus far, in robotics, rigid human models, which lack the accurate articulation and soft qualities humans exhibit in real life, are more commonly used \cite{8794160}.





\begin{table}[t]
\centering
\vspace{2mm}
\caption{Human Joint Limits} 
\label{tab:joint_limits}
\vspace{2mm}
\begin{tabular}{ccccc}
    \toprule
     & \textbf{Shoulder} &  & \textbf{Hip}  \\
    \midrule
    Flexion / Extension &  $160\degree$ / $49\degree$ &  & $110\degree$ / $14\degree$  \\
    Abduction / Adduction & $174\degree$ / $0\degree$ & & $33\degree$ / $22\degree$\\
    Medial / Lateral Rotation & $63\degree$ / $92\degree$ && $26\degree$ / $26\degree$ \\
    \midrule
    & \textbf{Elbow} &  & \textbf{Knee}  \\
    \midrule
    Flexion / Extension & $136\degree$ / $0\degree$ && $125\degree$ / $0\degree$ \\
    Pronation / Supination & $70\degree$ / $77\degree$ &&\\
    \midrule
    & \textbf{Lumb/Thor Spine} &  &   \\
    \midrule
    Flexion / Extension & $77\degree$ / $25\degree$ &&   \\
    Lateral Flexion (L / R)& $33\degree$ / $36\degree$ &&   \\
    Left / Right Axial Rotation & $46\degree$ / $46\degree$ &&  \\
    \toprule
    \end{tabular}\\
    Ranges are chosen from \cite{reese2016joint} and \cite{narimani2018three} and are for individuals aged 25--39 years. These measurements are taken via goniometer measurements from a neutral starting position.
\vspace{-3mm}
\end{table}

\begin{table}[t]
\centering
\caption{Joint Stiffness and Damping} 
\label{tab:stiffness_table}
\begin{tabular}{cc}
    \toprule
    \textbf{Lumbar and Thorasic Spine} & K   \\
    \midrule
    Flexion/Extension & $110\;N/mm$   \\
    Lateral Flexion/Extension& $110\;N/mm$   \\
    Axial Rotation & $1240\;N/mm$  \\
    \toprule
    \textbf{Other Joints} &  \\
    \midrule
    Spring Constant (K) & 0.072 Ncm/\textsuperscript{o}  \\
    Damping Coefficient (B) & [0.30, 0.55]\\
    \toprule
    \end{tabular}\\
    Spinal values are taken from \cite{panjabi1976three}. General joint spring constant values are calculated in \cite{peiros2025haptic} using data from \cite{KOLZ2015940,kuechle1997shoulder} and damping is from \cite{engin1984damping}.
\vspace{-6mm}
\end{table}

\section{Methods}




\subsection{Problem Statement}
\label{sec: probStatement}

A person is found in configuration $T_1$ (prone) and must be moved to goal configuration $T_2$ (supine) while satisfying the safety constraint $G(\boldsymbol{f},\boldsymbol{\theta}) \le 0$, where $\boldsymbol{f} \in \mathbb{R}^{n_c \times 3}$ collects the contact forces at $n_c$ contact points and $\boldsymbol{\theta}$ is the full human joint configuration.
$\boldsymbol{\theta}$ are bounded by the joint angles ranges in Table~\ref{tab:joint_limits} and the physics of Table~\ref{tab:stiffness_table}.
Critically, spinal torsion must remain as minimal as possible, as the literature does not define any deviation as acceptable. We will choose a maximum deviation of $10\degree$ in all spinal rotational directions.

For this formulation of human motion, reinforcement learning was chosen to learn the appropriate rolling strategy.
While optimal control offers formal guarantees, it would require differentiating through multi-body, intermittent-contact dynamics that change topology as ground and intra-body contacts form and break during rolling.
RL sidesteps this requirement by treating the physics simulator as a black-box transition model.
The problem is formulated as an MDP $\langle \altmathcal{S}, \altmathcal{C}, R, M \rangle$, where $\altmathcal{S}$ is the state space, $\altmathcal{C}$ is the control (action) space, $R$ is the reward function, and $M$ is the motion (transition) model.
We use Proximal Policy Optimization (PPO) \cite{schulman2017proximal} with 4,096 parallel environments, training for 6000 iterations with an actor--critic architecture comprising of hidden layers of 256, 256, and 128 units each with ELU activations, log standard deviation clamped to [0.05, 0.5], and learning-rate adapted to keep the KL divergence near 0.01.


\subsection{State Space $S$}
The observation vector $\boldsymbol{o}_t \in \mathbb{R}^{24}$ captures the minimal information the policy needs to decide the next force command.
Rather than exposing the full articulation state (which includes dozens of joint angles), we provide a compact representation centered on the torso orientation and the accumulated force history:
\begin{equation}
\label{eq:obs}
\boldsymbol{o}_t = \big[\,
\boldsymbol{q}_t,\;
c_t,\; p_t,\; \Delta p_t,\;
\tau_t,\;
H_{t},\; V_{t},\;
F,\;
\boldsymbol{e},
m,
h
\,\big]
\end{equation}
where $\boldsymbol{q}_t \in \mathbb{R}^4$ is the root-link quaternion (frame on the torso); $c_t \in [-1,1]$ is the \emph{closeness} metric; $p_t = (1 - c_t)/2 \in [0,1]$ is flip progress; $\Delta p_t = p_t - p_{t-1}$ is the progress rate; $\tau_t = t / T \in [0,1]$ is the normalized episode time; $H_{t}, V_{t} \in R^{4xT_n}$ are matrices of the horizontal and vertical force magnitudes at the contact points at each time step; $F$ is the total force magnitude summation of both contact points; $\boldsymbol{e}$ is vector of one-hot encoding of the selected contact bodies where $e_i \in \{0,1\}^K$, $m$ is the total mass of the model in kg and $h$ is the total height in m. The subscript $t$ always refers to a time step, $T$ is the total time, $T_n$ is the number of time steps, and $i$ is the index for the relevant body.

\subsection{Control Space $C$} 


The policy outputs a $21$-dimensional action $\boldsymbol{u}_t \in [-1,1]^{21}$ made up of vectors grouped as:
\vspace{-4mm}

\begin{equation}
\label{eq:action}
\boldsymbol{u}_t = \big[\,
\underbrace{\boldsymbol{\ell}_0, \; \boldsymbol{\ell}_1}_{\text{body selection}},
\;
\underbrace{\boldsymbol{a}_0,\; \boldsymbol{a}_1}_{\text{application pts}},\;
\underbrace{\boldsymbol{r}_0}_{\text{regrasp}}, \;
\underbrace{\delta h_0,\; \delta v_0,\; \delta h_1,\; \delta v_1}_{\text{force commands}}
\big]
\end{equation}


\begin{figure*}
\vspace{2mm}
    \centering
    \begin{subfigure}[t]{0.17\linewidth}
        \includegraphics[width=\linewidth]{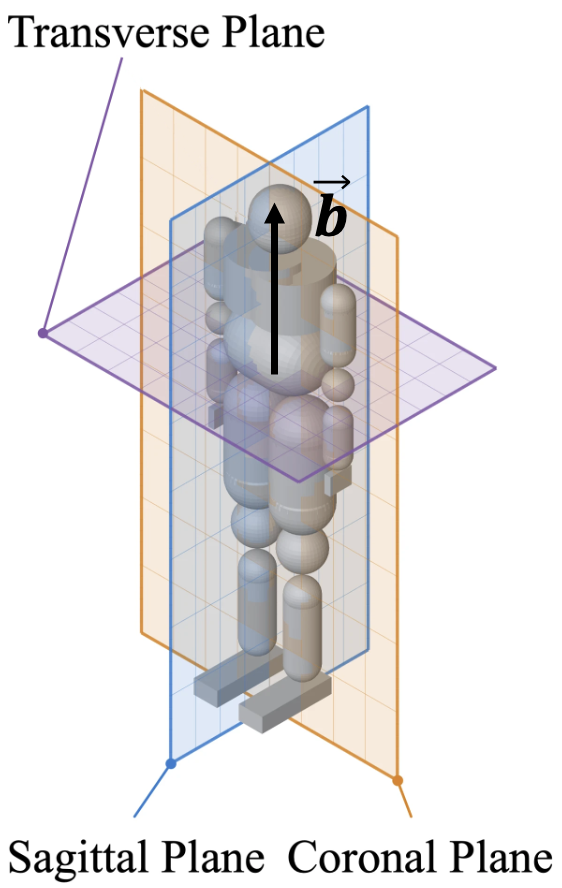}
        \caption{}
    \end{subfigure}
    \begin{subfigure}[t]{0.23\linewidth}
        \includegraphics[width=\linewidth]{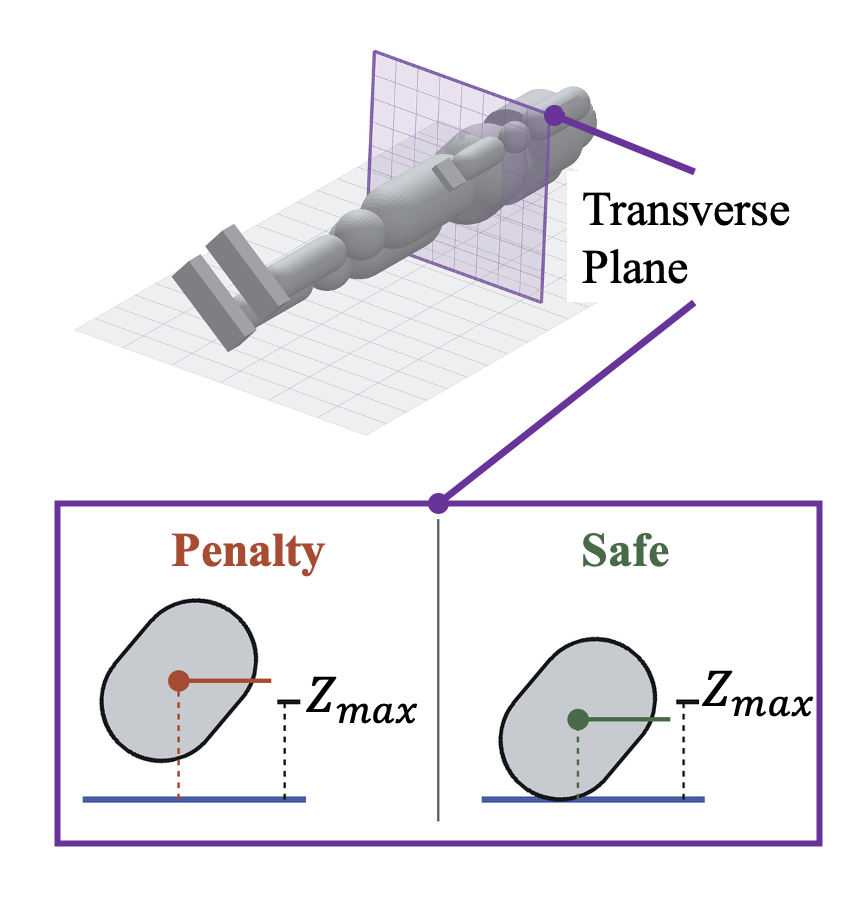}
        \caption{}
    \end{subfigure}
    \begin{subfigure}[t]{0.25\linewidth}
        \includegraphics[width=\linewidth]{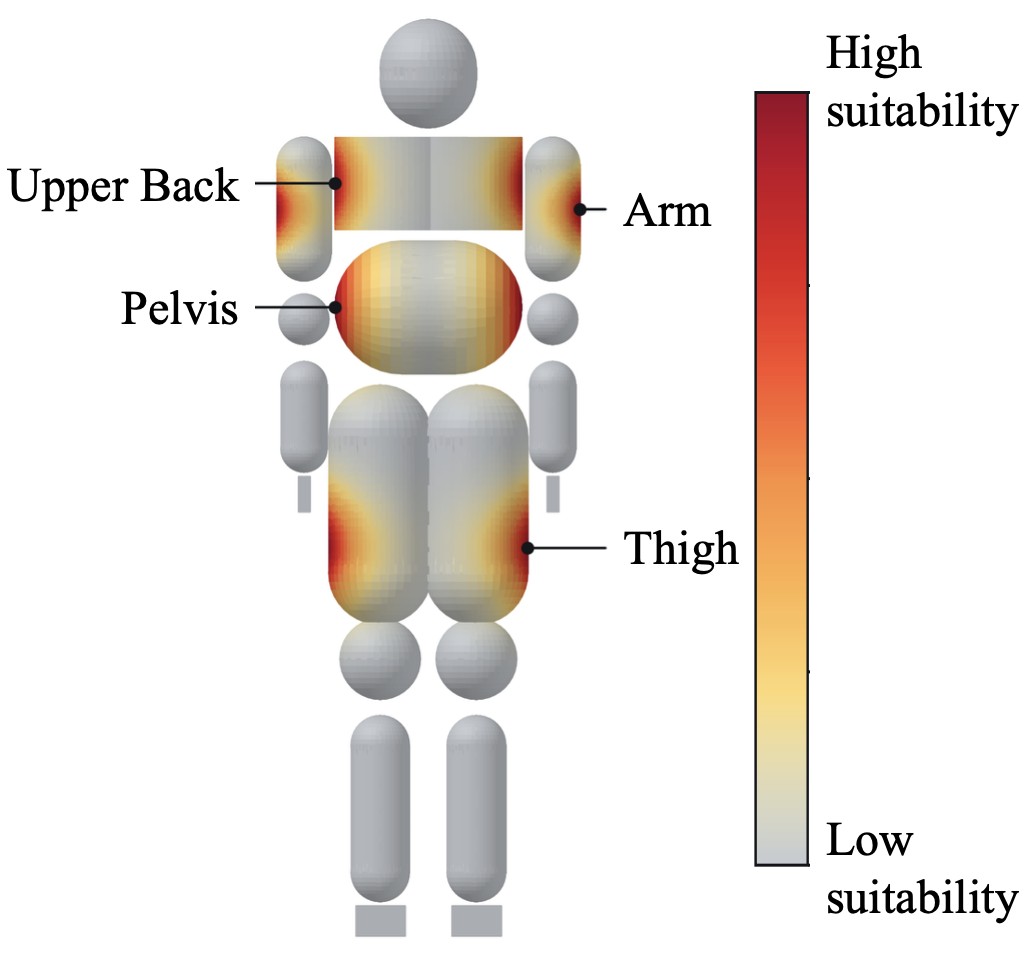}
        \caption{}
    \end{subfigure}
    \begin{subfigure}[t]{0.23\linewidth}
        \includegraphics[width=\linewidth]{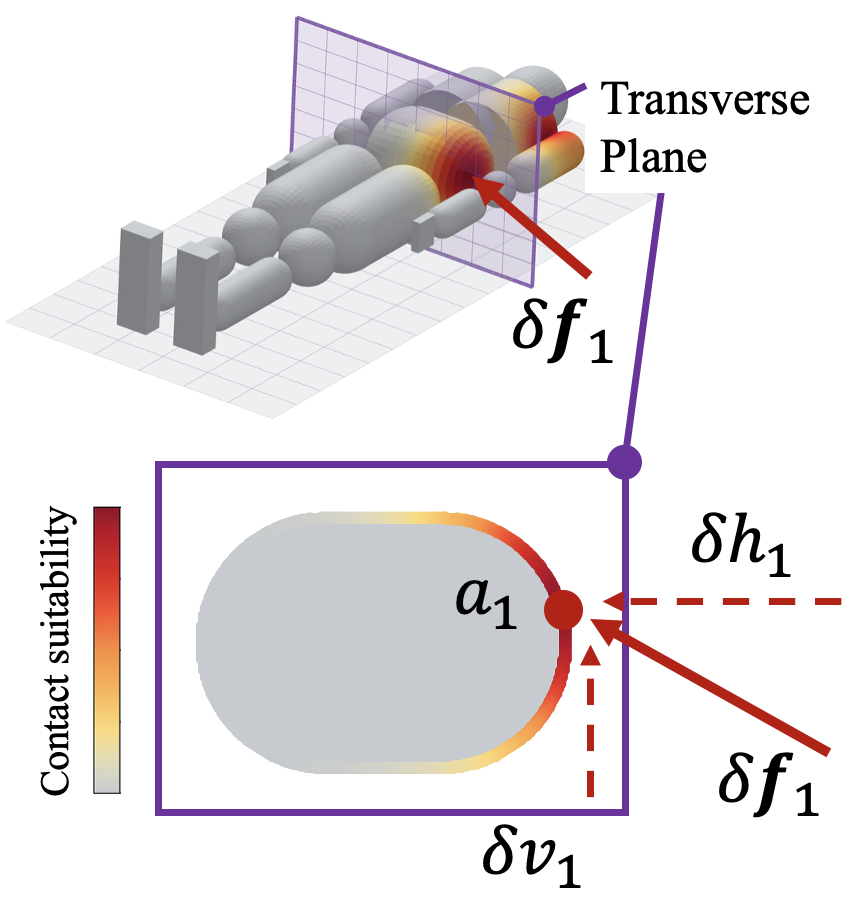}
        \caption{}
    \end{subfigure}
    \caption{(a) Biological frame nomenclature for planes of dissection; (b) visual definition for lifting penalty describing the geometric center height measurement relative to $z_{max}$, the height associated with losing ground contact, in the transverse plane; (c) Heat map depicting the contact regions included in contact point selection; dark coloring distinguishes the central region of the contact and the lighter colors depict the allowable region for motion during rolling; (d) Breakdown of the contact resultant force into its vertical, $\delta v_i$, and horizontal, $\delta h_i$, components in the body frame. In the world frame, the delta resultant force is $\delta \boldsymbol{f}_{i} = [\delta h_i^{\top}\boldsymbol{\hat{x}}, \delta h_i^{\top}\boldsymbol{\hat{y}}, \delta v_i]$. The force is applied at contact point $a_i$.}
    \vspace{-5mm}
    \label{fig:lift_penalty}
\end{figure*}

\subsubsection{Contact body selection}
At the start of each episode, the policy's logits determine the two contact bodies, which are fixed for the duration of the episode to encourage stable planning. The logit vectors $\boldsymbol{\ell}_i \in \mathbb{R}^K$ select which body link receives the force. $K=4$ candidates are chosen from; these regions appear highlighted in {Fig~\ref{fig:lift_penalty}(c)} as hot spots on the model. 

\subsubsection{Application Points} Each $\boldsymbol{a}_i \in [-1,1]^3$ specifies a point on the selected contact body in its local frame, scaled by the body's bounding dimensions as seen in {Fig~\ref{fig:lift_penalty}(d)}, labeled as $a_1$ with force $A$ acting on it. The application point starts as a fixed point on the body surface where contact would occur with the robot (or human). 

\subsubsection{Force Commands} The force deltas $\delta h_i, \delta v_i \in [-1,1]$ are presented as the horizontal and vertical impulse commanded at contact point $a_i$ seen in {Fig~\ref{fig:lift_penalty}(d)}. The maximum impulse per time step is mapped to $5N$ and minimum is $0N$.
The resulting impulse in world coordinates at contact $a_i$ is $\delta \boldsymbol{f}_{i} = [\delta h_i^{\top}\boldsymbol{\hat{x}}, \delta h_i^{\top}\boldsymbol{\hat{y}}, \delta v_i]$. Force is clamped to magnitudes of $[0,200 N]$ vertically and $[0,300 N]$ horizontally.

\subsubsection{Re-grasp Condition} $\boldsymbol{r}_0$ indicates whether the contact point jumps to its mirror for the descent portion of the roll. Typically, humans use both hands and will trade hands during the top portion of the roll to switch from accelerating to decelerating the body. This condition is allowed once progress has reached 40\% of the total roll and can only be used once during the duration of the roll.

\subsection{Reward Shaping $R$}

As with typical reward-based policies, the reward function couples task completion with task success, which in this case are defined through the safety considerations needed to maintain spinal stability. Each component is individually switchable, enabling the ablation study in Section~\ref{sec:ablation}. Common rewards include progress measures, a force magnitude penalty, and success bonuses. Episodes terminate due to time limits, safety violations, or task completion.

\begin{table}[t]
    \vspace{2mm}
    \centering
    \caption{Reward and Penalty Weights}
    \begin{tabular}{cccc}
    \toprule
    Weight Symbol & Magnitude & Description \\
    \toprule
    $- \alpha_{w}$ & 200 & Gentle Roll \\
    \hline
    $- \alpha_f$ & 5 & Force Penalty \\ 
    \hline
    $- \alpha_l$ & 1 & Lift Penalty \\
    \hline
    $- \alpha_{sp}$ & 50 & Spinal Alignment Penalty \\
    \hline
    $- \alpha_{ax}$ & 50 & Spin penalty \\
    \hline
    $\lambda_{o}$ & 50 & No Overshoot Bonus \\
    \hline
    $\lambda_s$ & 200 & Task Success \\
    \hline
    $-\lambda_{f}$ & 100 & Physical Failure \\
    \hline
    $-\lambda_{T}$ & 20 & Timeout \\
    \hline
    $-\alpha_{inj}$ & 3 & Injury Motion Prevention \\
    \bottomrule
    \end{tabular}
    \vspace{-4mm}
    \label{tab: weights}
\end{table}

\subsubsection{Gentle Roll}
To encourage a gentle roll, the policy is penalized for going above a safe desirable speed. The speed cap was chosen to be $\Delta p_{max} =.65 \frac{rad}{s}$:

\begin{equation}
    r_w = \sum_i
    \begin{cases}
    0  & \Delta p \leq \Delta p_{max}\\
    (\Delta p - \Delta p_{max}) & \Delta p > \Delta p_{max}\\
\end{cases}
\end{equation}

where $\Delta p$ measures angular progress and is an equivalent measure to angular velocity. 

\subsubsection{Lift Penalty} An ideal rolling motion maintains continuous contact with the ground throughout the roll. To avoid the challenge of tracking the contact points with the ground, which are instantaneous and constantly redefined
a proxy measurement was defined: the height of the center of geometry relative to the ground plane. The assumed maximum height is taken as the radius of the individual unique USD model torso and is used to set $z_{max}$: 

\begin{equation}
r_{\text{l}} =  \sum_i
\begin{cases}
0  & z \leq z_{max}\\
\boldsymbol{f}_{i,t}^\top\hat{\boldsymbol{n}} & z > z_{max}\\ 
\end{cases}
\label{eq:r_lift}
\end{equation}

where the piecewise function defines a penalty only if the body has lifted off the ground (Fig~\ref{fig:lift_penalty}(b)). In this state, the forces in the +z direction are penalized proportionally to the lift they generate on the model. This reduces throwing in favor of rolling the body and minimizes improper task completion, such as rolling over the head or shoulder rather than the full side profile. 

\subsubsection{Total Spinal Alignment Penalty} 
\label{Sec: alignment}

As described in section \ref{sec: background}, spinal stability is a key factor in the success of the roll. Specifically, maintaining spinal alignment throughout the roll sequence. In the reward function, a penalty is included based on the deviation of spine. In our model the spine is made of 3 revolute joints each having a joint limit [$\alpha_{max}$, $\beta_{max}$, $\gamma_{max}$]. The deviation from zero as a percentage of its maximum range squared is what is penalized:
\vspace{-2mm}
\begin{equation}
    r_{\text{sp}} = \big(\frac{\alpha}{\alpha_{max}}\big)^2 
    + \big(\frac{\beta}{\beta_{max}}\big)^2 
    + \big(\frac{\gamma}{\gamma_{max}}\big)^2 
\end{equation}

where [$\alpha$, $\beta$, $\gamma$] are the angles, of the spine, between the torso components. This directly penalizes flexion/extension, lateral flexion/extension, and axial rotation during the roll, enforcing the clinical requirement that the head and upper torso move with the hips throughout the maneuver, minimizing spinal deflection. For human legibility and relevance to clinical markers, this will be shown as a deviation angle $\phi$, derived from the angle-axis rotation formulation calculated from our sequence of Euler angles [$\alpha$, $\beta$, $\gamma$]. The conservative limit for the deviation is $16.79\degree$, equivalent to each angle reaching $10\degree$, as described in section~\ref{sec: probStatement}.

\subsubsection{Shoulder-Hip Motion Disagreement Penalty} In the transverse plane, progress is measured by how much the body moves toward the goal angle. Both the shoulders and hips should progress together. To ensure the entire body rolls as a coordinated unit and makes progress, the relative progress is incentivized to move together. This is done by comparing the progress of the upper and lower body separately and comparing them to minimize the difference in  
\vspace{-2mm}
\begin{equation}
r_{\text{dis}} = \big( | p_{0,t} - p_{1,t} |\big)
\label{eq:r_disagree}
\end{equation}

This term prevents shoulder rolling while leaving the pelvis behind, enforcing a whole-body log roll.

\subsubsection{Spin-axis Alignment Penalty} The rotation of the body about the normal to the ground plane is defined by the vector along the midline cross section of the sagittal and coronal plane shown in Fig \ref{fig:lift_penalty}(a) as $\vec{b}$, going from pelvis to head at the starting pose and its current position. The penalty uses the angle between the current and starting orientation.
\vspace{-2mm}
\begin{equation}
r_{\text{ax}} = \, (1 - \abs{\cos\theta_{\text{ax}}})
\label{eq:r_axis}
\end{equation}

where $\theta_{\text{ax}}$ is the angle between the current and initial directions. This penalty also encourages balanced forces at both contact points, reducing excessive rotation. 

\subsection{Injuries}

\subsubsection{Observation Additions}

The observation vector $\boldsymbol{o}_t \in \mathbb{R}^{26}$ additionally captures which joint is injured and the relative motion of that specified joint.
\begin{equation}
\label{eq:obs}
\boldsymbol{o}_t = \big[\,
\boldsymbol{q}_t,\;
c_t,\; p_t,\; \Delta p_t,\;
\tau_t,\;
H_{t},\; V_{t},\;
F,\;
\boldsymbol{e},
m,
h
, j_i,\; \dot{\boldsymbol{q}}\big]
\end{equation}

where $j_i$ is the injured joint and $\dot{\boldsymbol{q}}$ is Euclidean norm of the injured joints' velocities. In this case, it is assumed the injury is known and is not discovered. 

\subsubsection{Controls Additions}

The control space is not augmented for the inclusion of injury consideration in the policy.


\subsubsection{Rewards Additions}
This reward function couples the previous task of spinal stability with minimizing motion in areas marked as injured. The additional reward $r_{\text{inj}}$ is defined as:
\vspace{-2mm}
\vspace{-.5mm}
\begin{equation}
    r_{\text{inj}} = \, \frac{\Delta t}{T} \cdot \text{clip}(\frac{\left\|\boldsymbol{q}_{dif} \right\|}{.15}, 0, 1)
\end{equation}

\subsection{Domain Randomization}

\subsubsection{Human Body Model}
\label{sec: human_model_torso}

The human model requires specific articulations and modeling elements to understand the algorithm's performance and execution relative to real-world humans. In this work, a base human model is used, comprising a total of 21 degrees of freedom: 5 per arm [shoulder flexion, shoulder abduction, humeral rotation, elbow flexion, ulnar rotation], 4 per leg [hip flexion, hip abduction, femoral rotation, knee flexion], 3 mid-back rotations [spinal flexion, lateral spinal flexion, axial rotation]. 

The human model uses a rigid head-and-upper-torso frame as its center. The right and left arms attach to the torso, and the pelvis sits at the base. The pelvis is connected to the legs. All of the joint limits can be found in the Table \ref{tab:joint_limits}. For proper dynamic properties, particularly the spine, the stiffness in each anatomical degree of freedom was found along with an approximate damping coefficient. A general tendon stiffness was also found for the joints. Table \ref{tab:stiffness_table} lists these values.

\subsubsection{Randomized Domains}
\begin{figure}[t]
\vspace{2mm}
    \centering
    \includegraphics[width=0.4\linewidth]{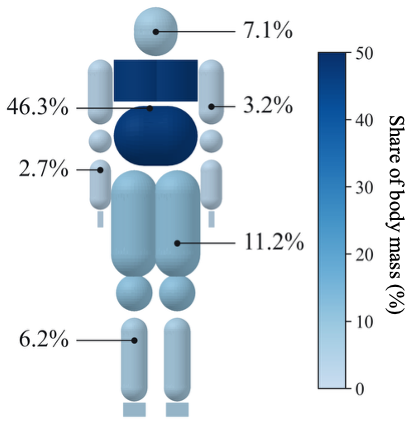}
    \caption{Average weight distribution in a human used for domain randomization of weight~\cite{clauser1969weight}.}
    \label{fig:weight_distrib}
    \vspace{-3mm}
\end{figure}

For training, the initial URDF skeleton was scaled to different heights evenly distributed throughout the range [1.4 m, 2 m], which corresponds to the bottom 1st percentile female and the top 1st percentile male, respectively, for 20-year-olds \cite{WHO2006ChildGrowth, Fryar2016Anthropometric, NCDRisC2020HeightBMI}. The masses, COMs, and moments of inertia are varied with reference to the data in \cite{clauser1969weight}, shown visually in {Fig~\ref{fig:weight_distrib}}. Typical friction randomization was also used.

\subsubsection{Injury}
Injury was prescribed in 3 categories: (1) a full human model with a designated injury location (2) a partial amputee where the lower portion of the limb was removed and the upper limb joint was considered for the injury metric (3) full amputee where there is no additional metric to be measured as the joint is fully removed, but the default safety measures and success remain.

\section{Experiments and Results}


\subsection{Human Deployment Models}
\label{sec: human_model_torso_deploy}

\begin{figure}[t]
\vspace{2mm}
    \centering
    \includegraphics[width=\linewidth]{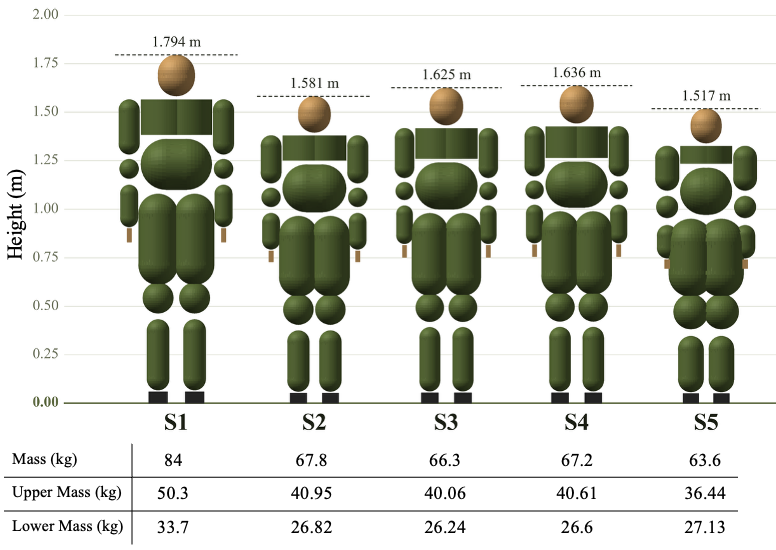}
    \caption{Heights, masses, and limb scaling of the five human models used for deployment.}
    \label{fig:human_models}
    \vspace{-3mm}
\end{figure}

The 5 human models used for testing (Fig.~\ref{fig:human_models}) are augmented based on circumferential measurements of the arms and waist for more accurate individualized weight distribution~\cite{lee2023contactless}. The policies developed in this work are deployed on these models for evaluation. These models provide realistic human parameters without safety risk to real-world humans. The specific weight distributions and heights, moments of inertia, and length distributions of the 5 subject models (Fig. \ref{fig:human_models}) are not used in training.

\begin{figure*}[t!]
\vspace{2mm}
    \centering
    \setlength{\tabcolsep}{2pt}
    \begin{tabular}{>{\centering\arraybackslash}m{1em} *{5}{m{0.18\textwidth}}}
        \rotatebox{90}{\textbf{Ours}} &
        \includegraphics[trim= 2cm 0cm 4cm 0cm, clip, width=0.18\textwidth]{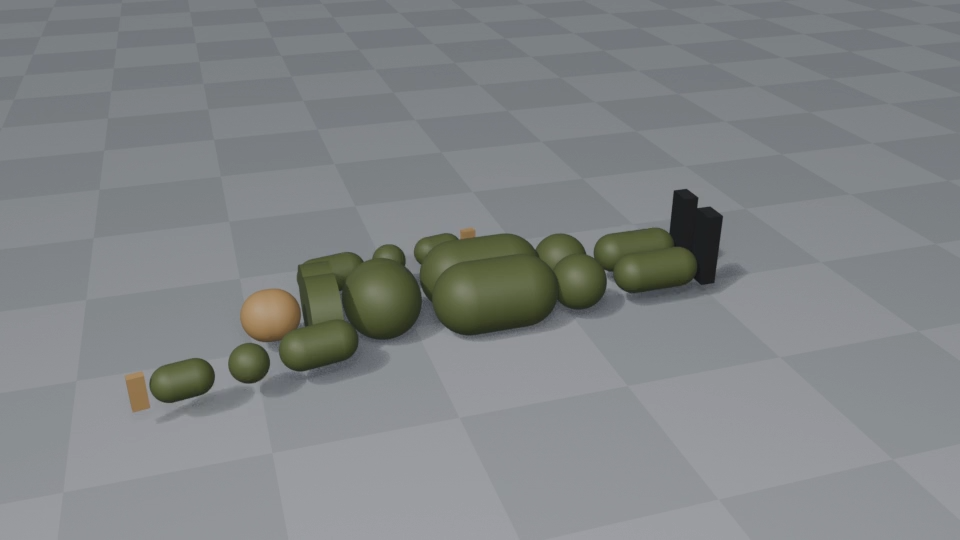} &
        \includegraphics[trim= 2cm 0cm 4cm 0cm, clip, width=0.18\textwidth]{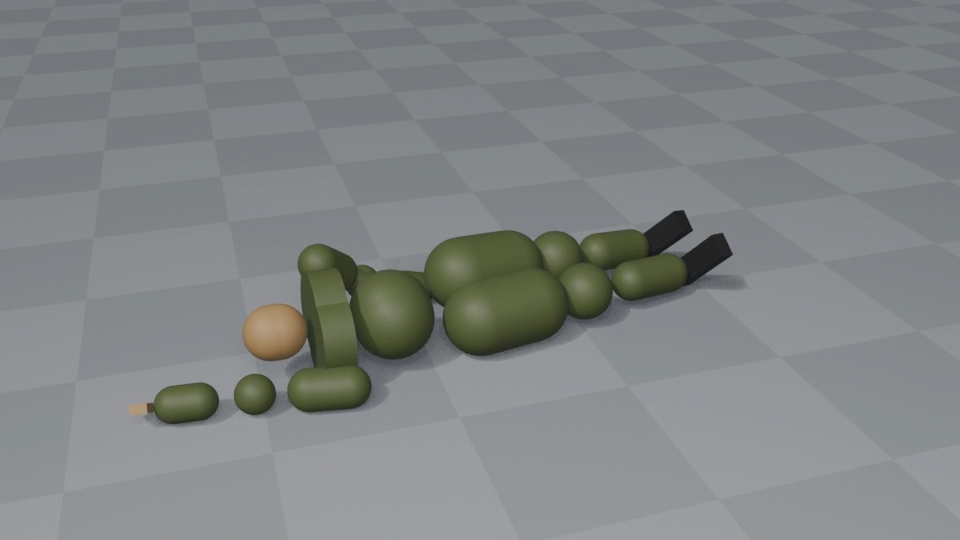} &
        \includegraphics[trim= 2cm 0cm 4cm 0cm, clip, width=0.18\textwidth]{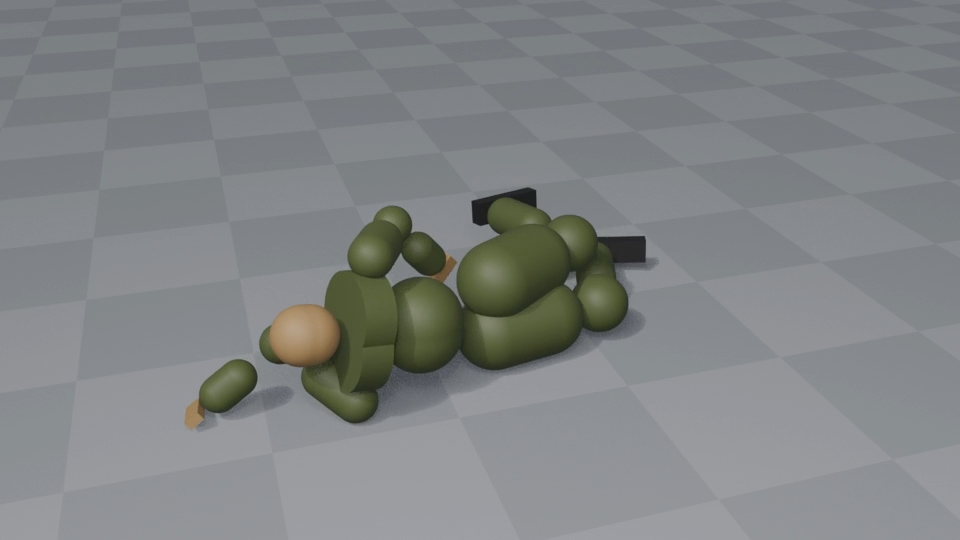} &
        \includegraphics[trim= 2cm 0cm 4cm 0cm, clip, width=0.18\textwidth]{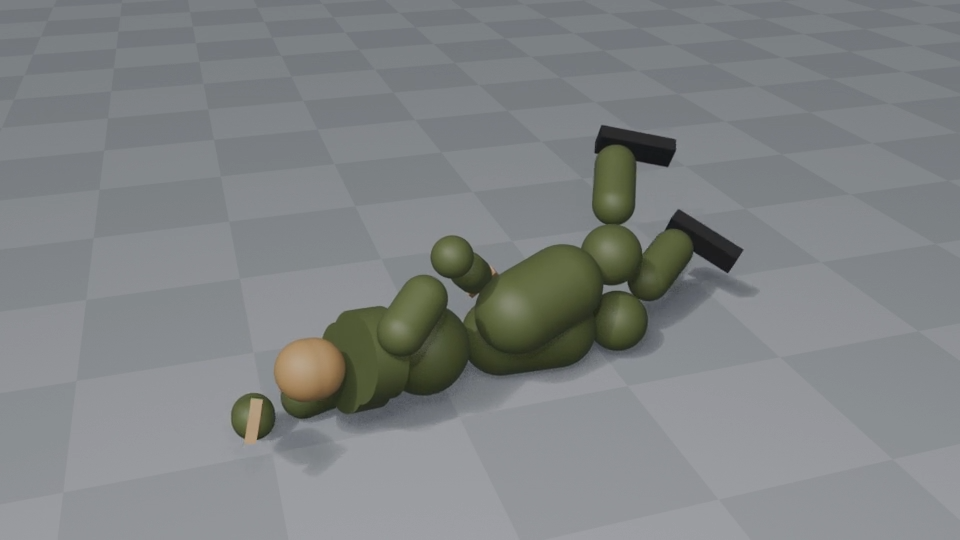} &
        \includegraphics[trim= 2cm 0cm 4cm 0cm, clip, width=0.18\textwidth]{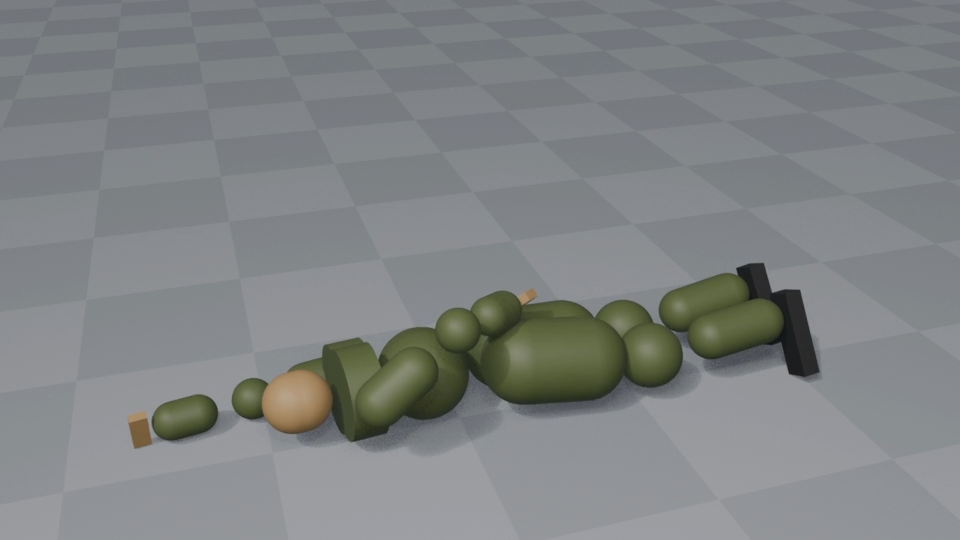} \\

        \rotatebox{90}{\textbf{Uniform}} &
        \includegraphics[trim= 2cm 0cm 4cm 5.2cm, clip, width=0.18\textwidth]{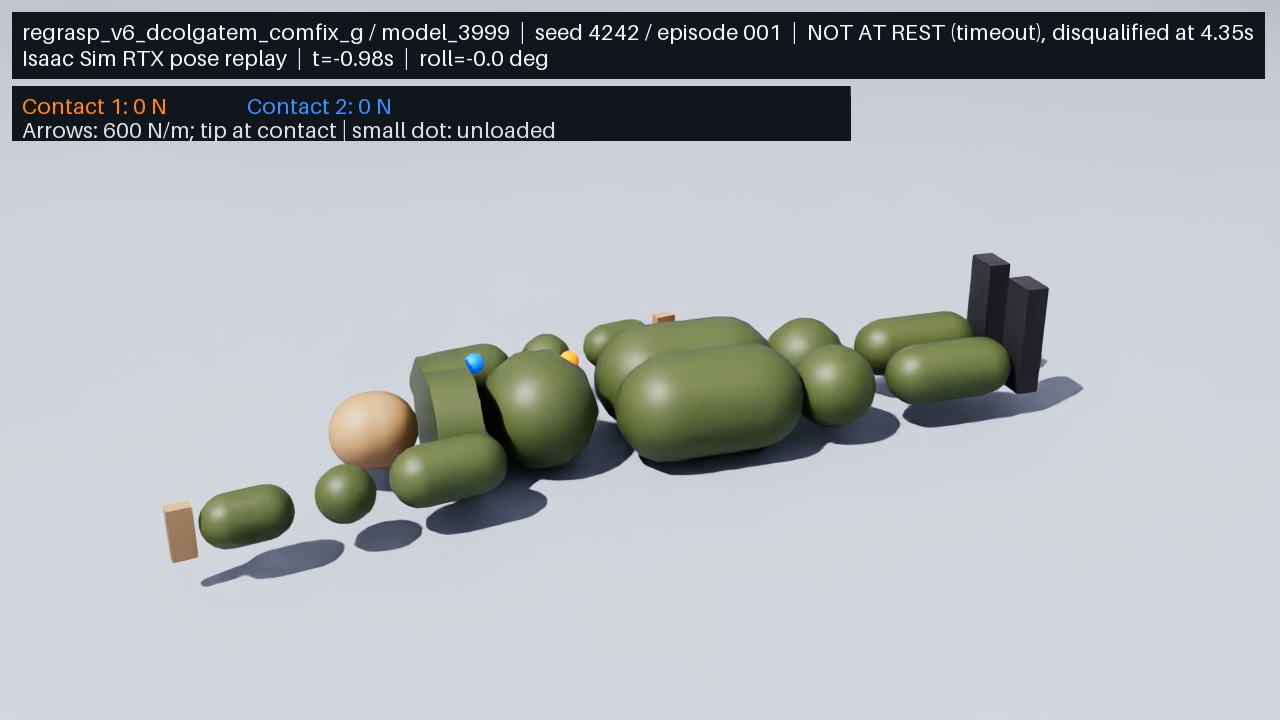} &
        \includegraphics[trim= 2cm 0cm 4cm 5.2cm, clip, width=0.18\textwidth]{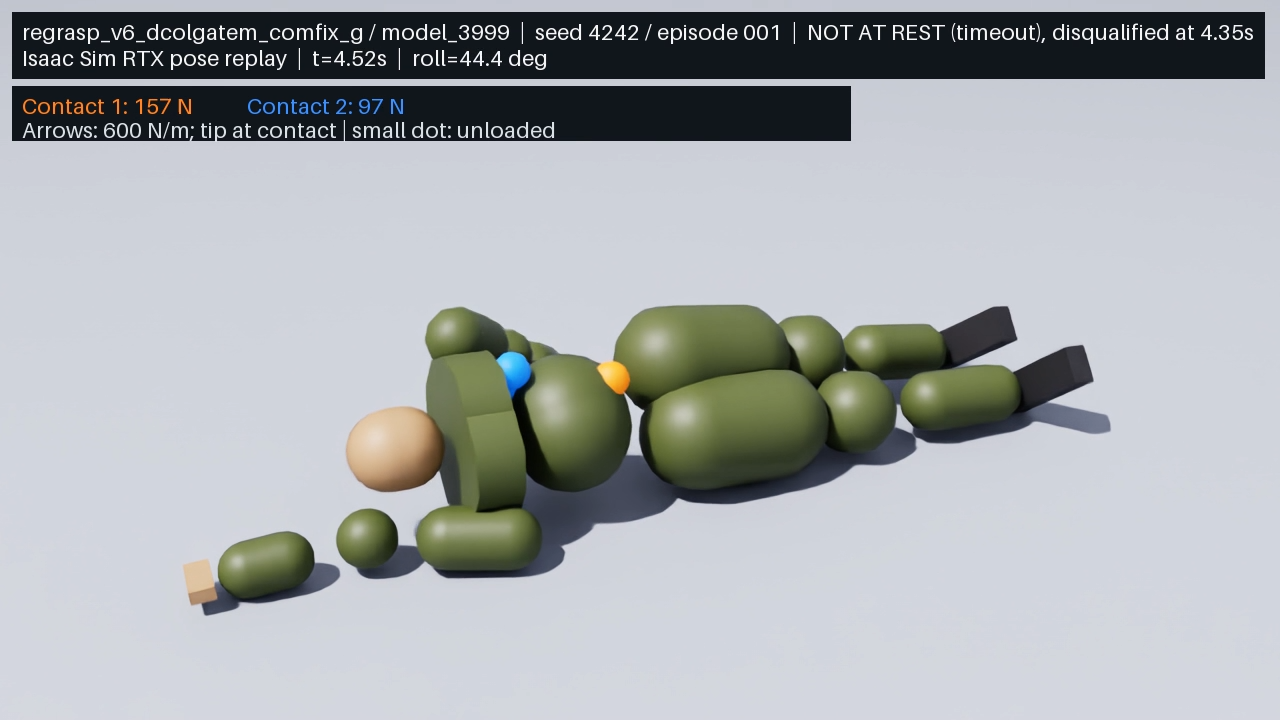} &
        \includegraphics[trim= 2cm 0cm 4cm 5.2cm, clip, width=0.18\textwidth]{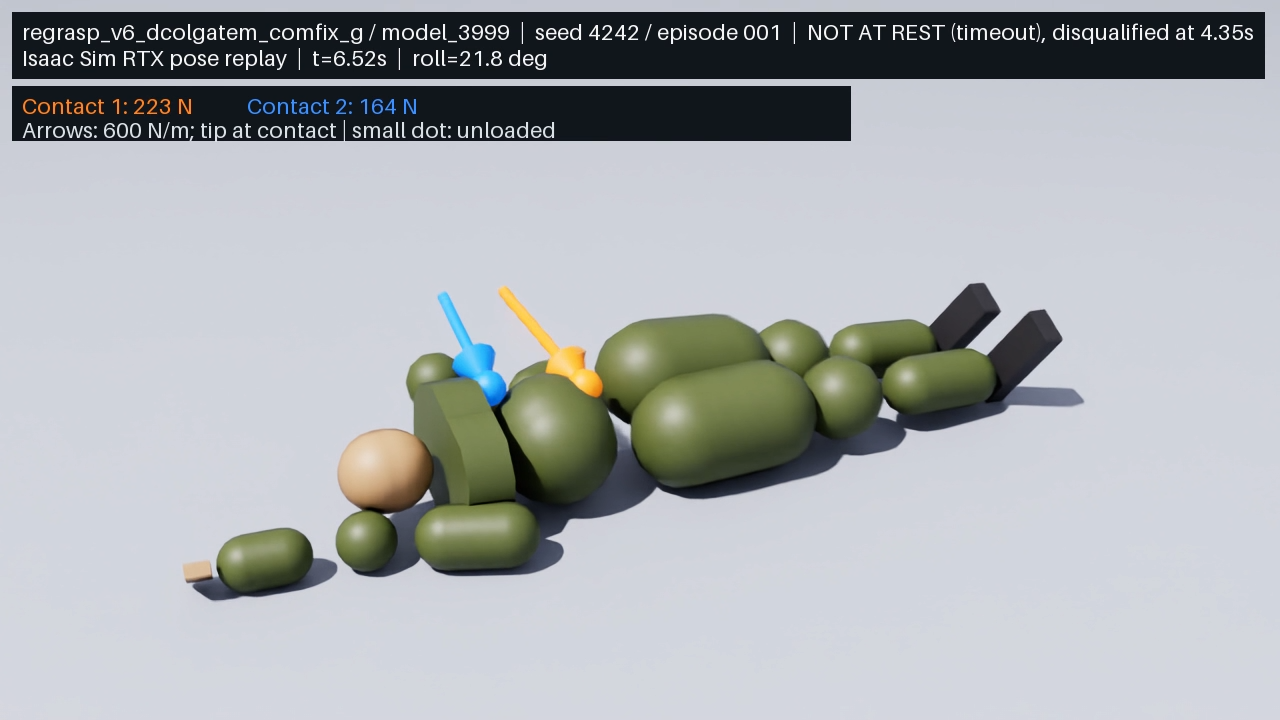} &
        \includegraphics[trim= 2cm 0cm 4cm 5.2cm, clip, width=0.18\textwidth]{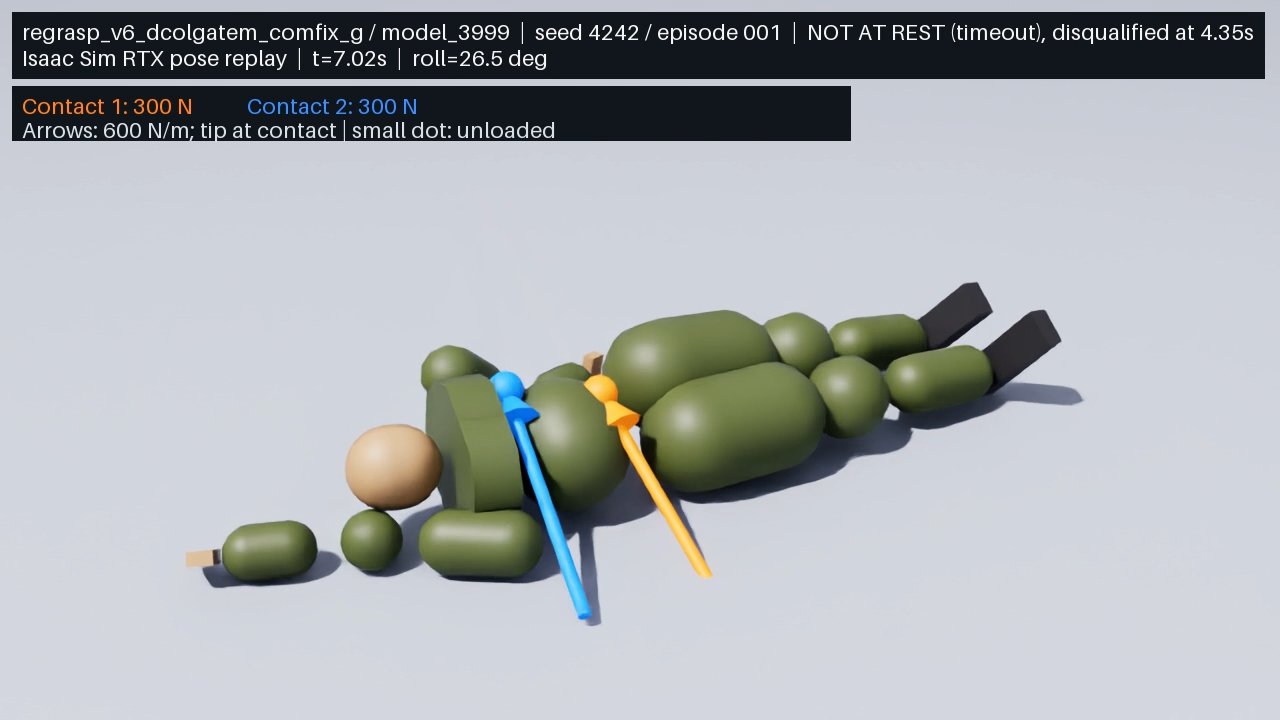} &
        \includegraphics[trim= 2cm 0cm 4cm 5.2cm, clip, width=0.18\textwidth]{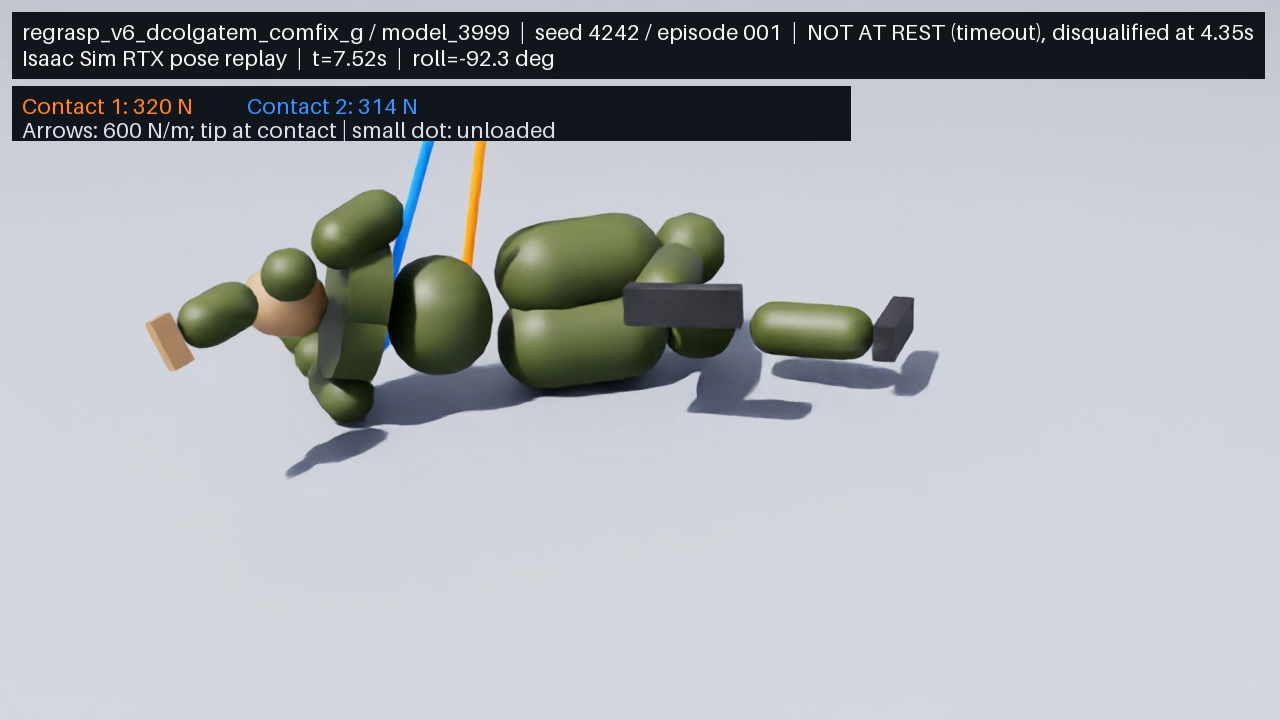} \\

        \rotatebox{90}{\textbf{FF 95th}} &
        \includegraphics[trim= 2cm 0cm 4cm 3cm, clip, width=0.18\textwidth]{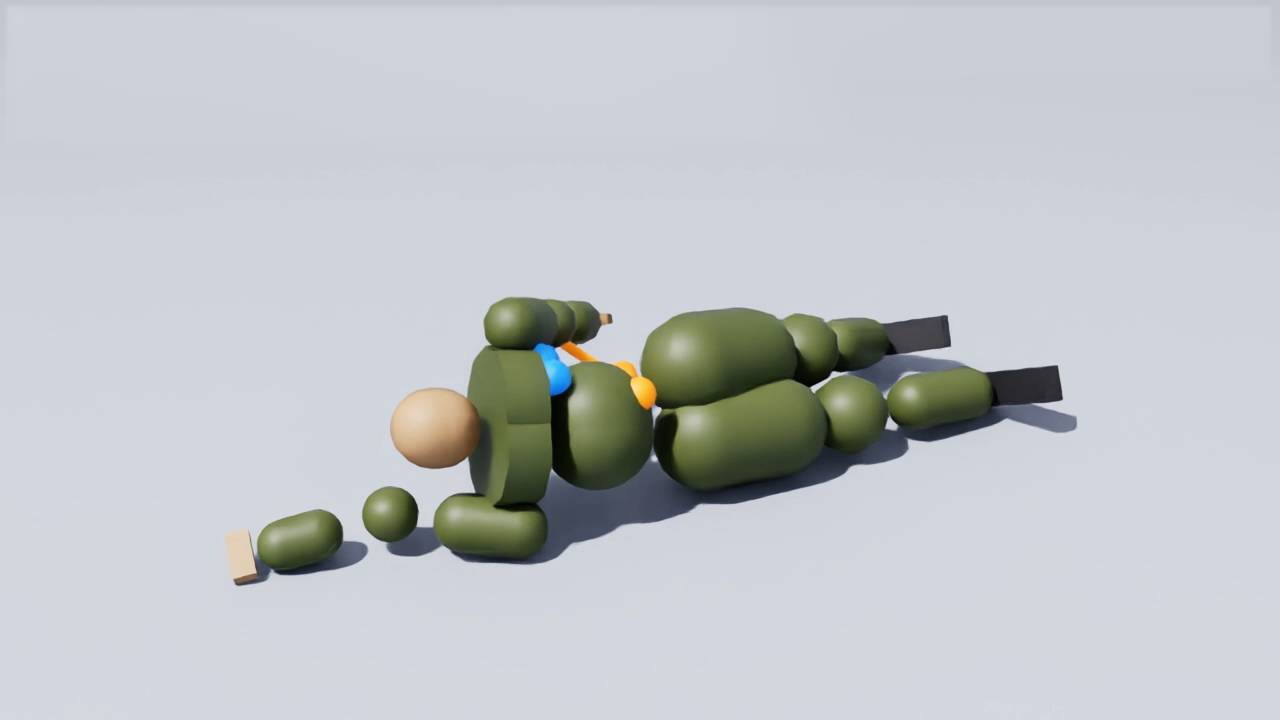} &
        \includegraphics[trim= 2cm 0cm 4cm 3cm, clip, width=0.18\textwidth]{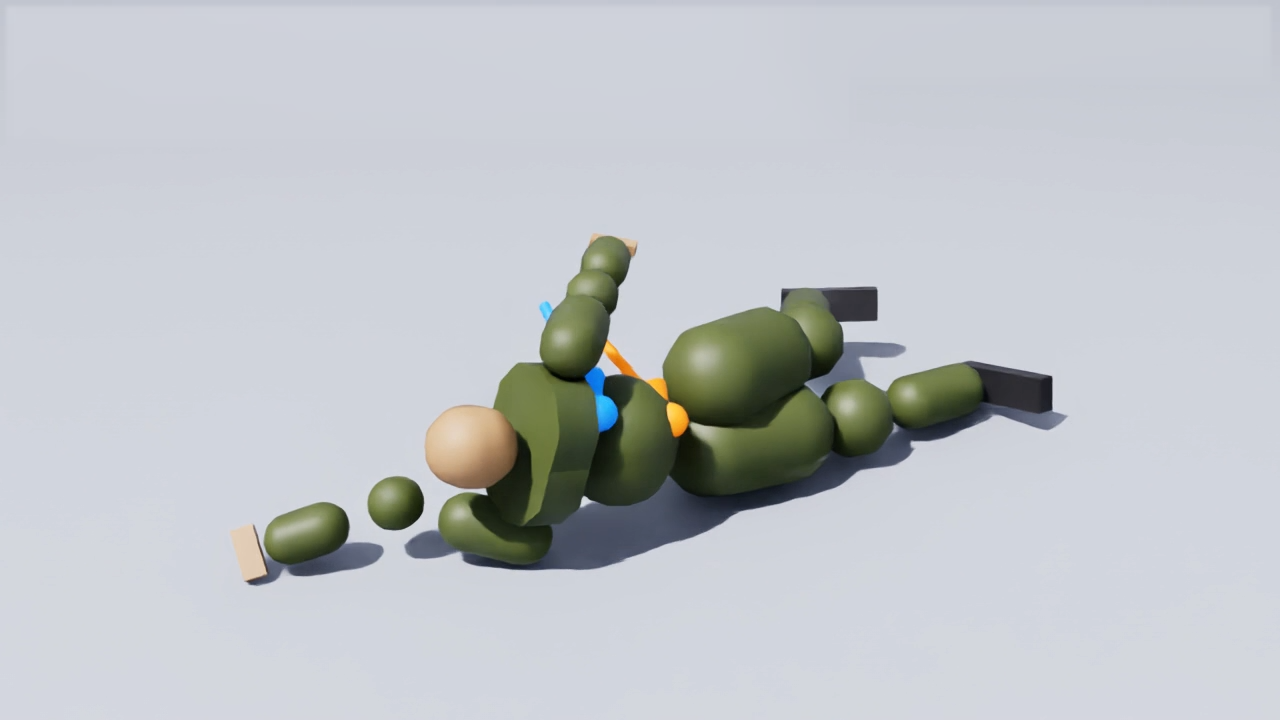} &
        \includegraphics[trim= 2cm 0cm 4cm 3cm, clip, width=0.18\textwidth]{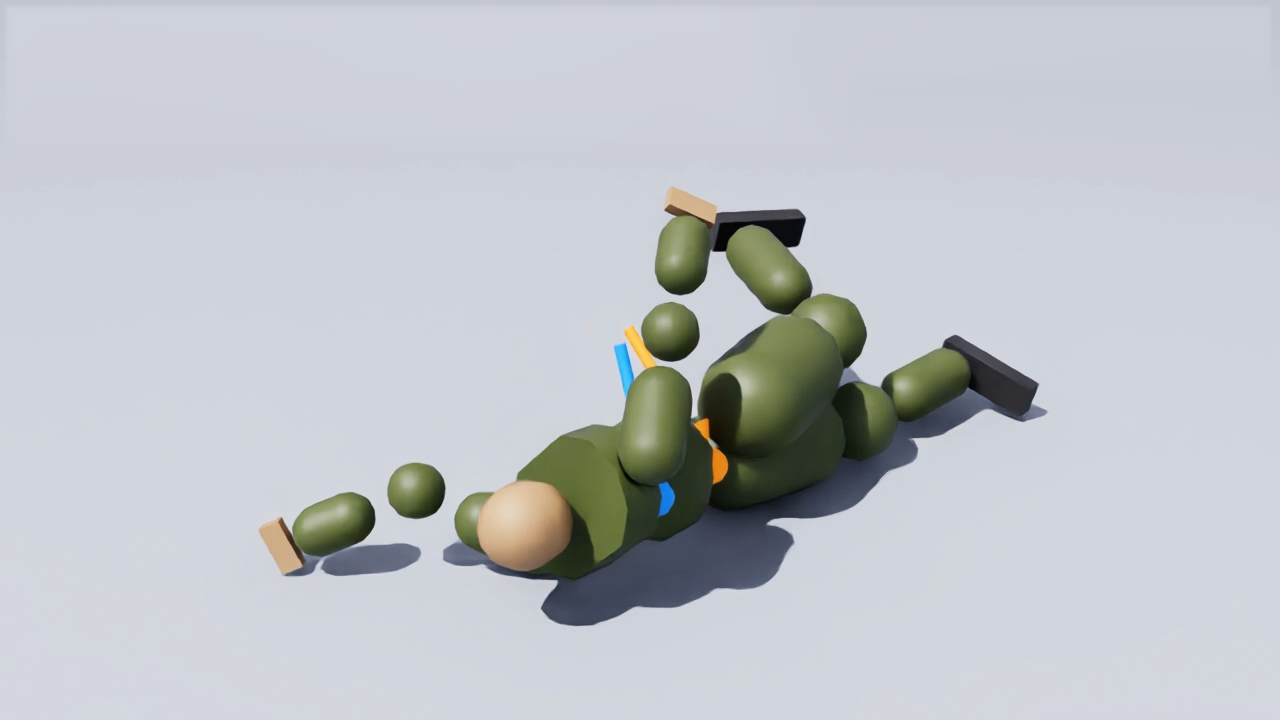} &
        \includegraphics[trim= 2cm 0cm 4cm 3cm, clip, width=0.18\textwidth]{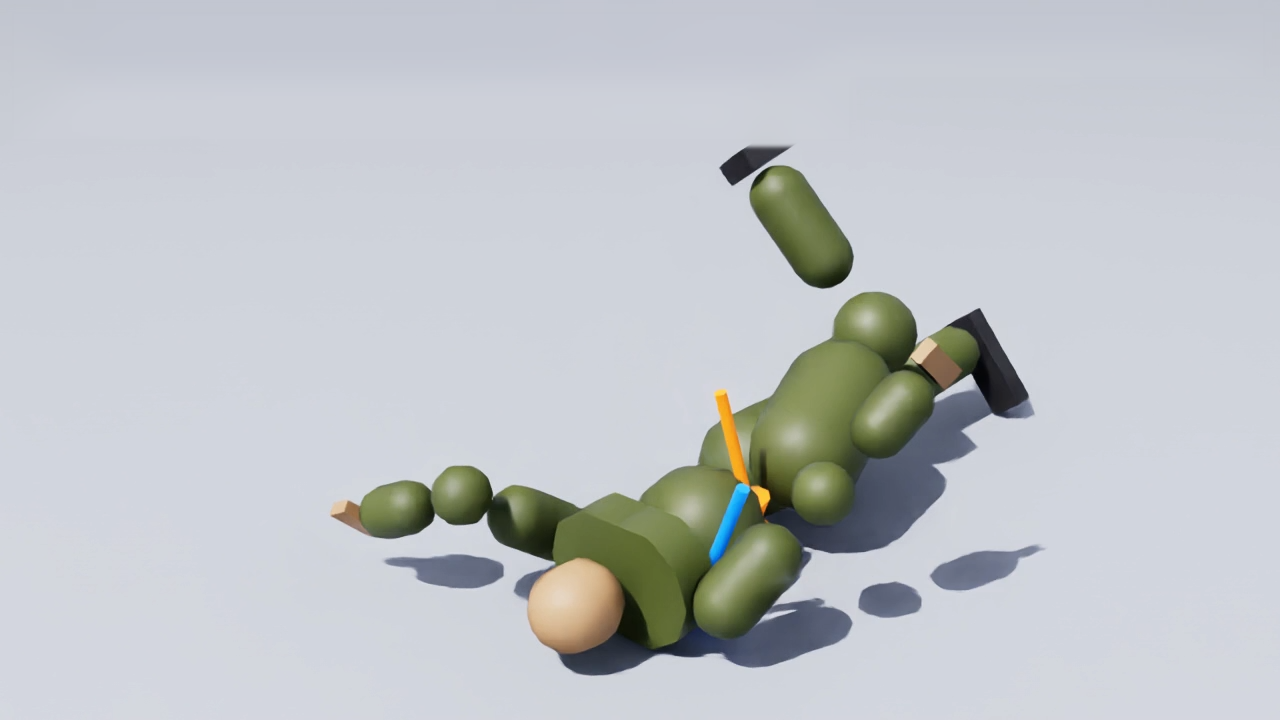} &
        \includegraphics[trim= 2cm 0cm 4cm 3cm, clip, width=0.18\textwidth]{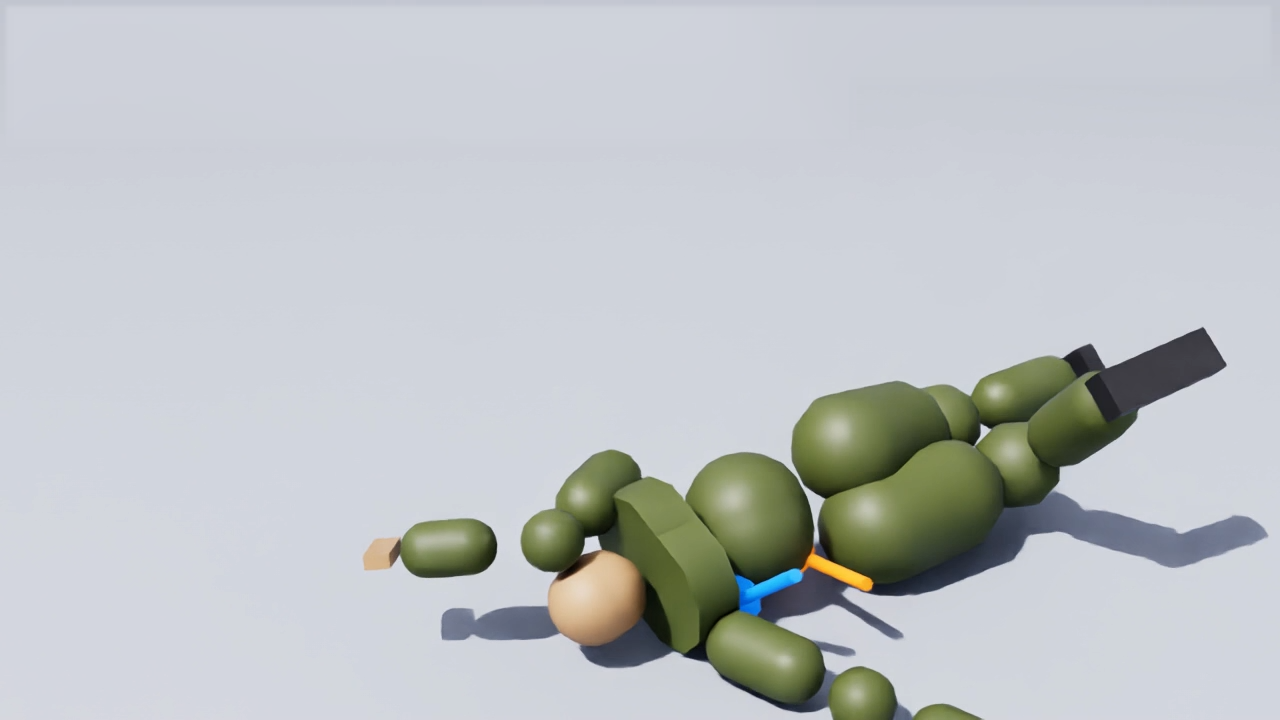} \\

        \rotatebox{90}{\textbf{Arm-L-2} }&
        \includegraphics[trim= 2cm 0cm 4cm 5.2cm, clip, width=0.18\textwidth]{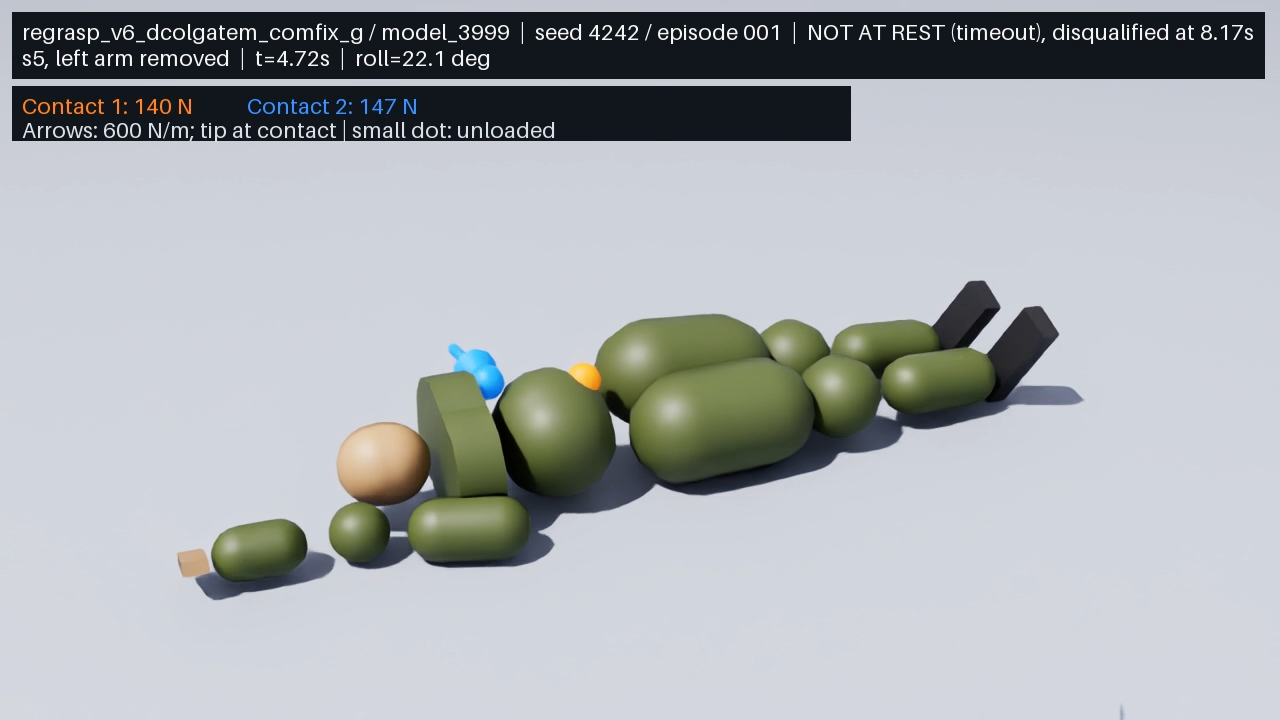} &
        \includegraphics[trim= 2cm 0cm 4cm 5.2cm, clip, width=0.18\textwidth]{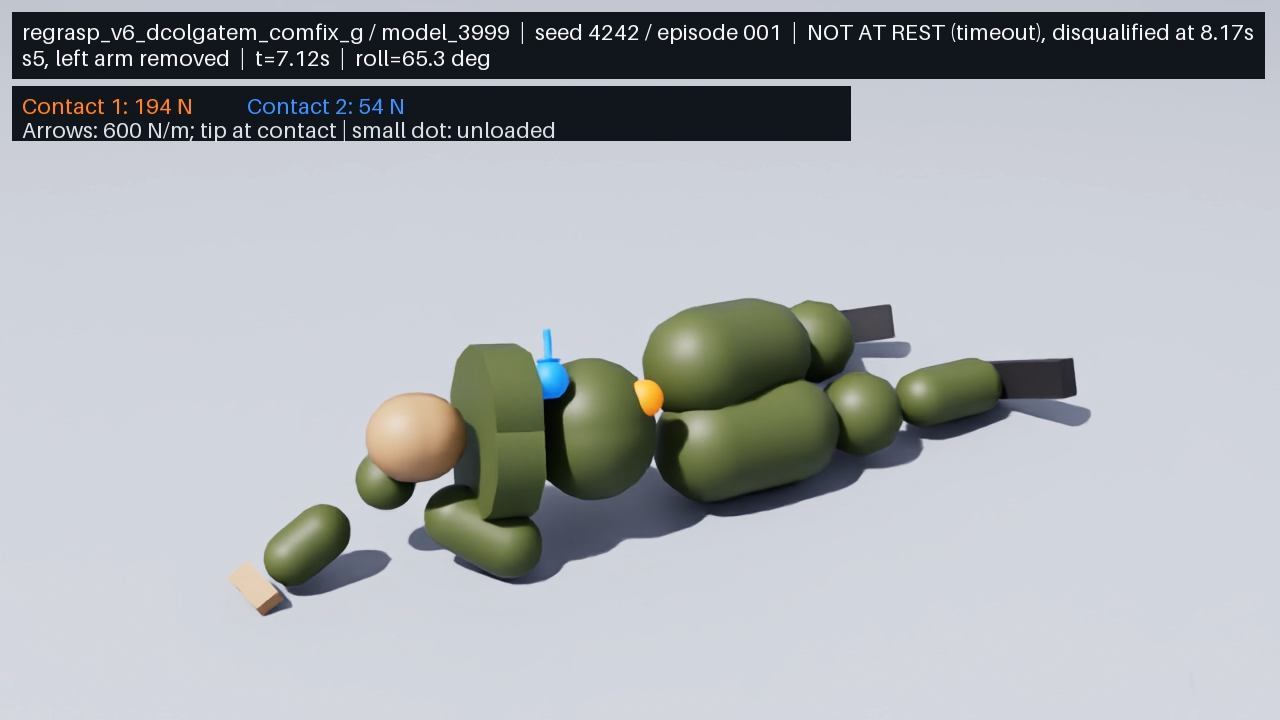} &
        \includegraphics[trim= 2cm 0cm 4cm 5.2cm, clip, width=0.18\textwidth]{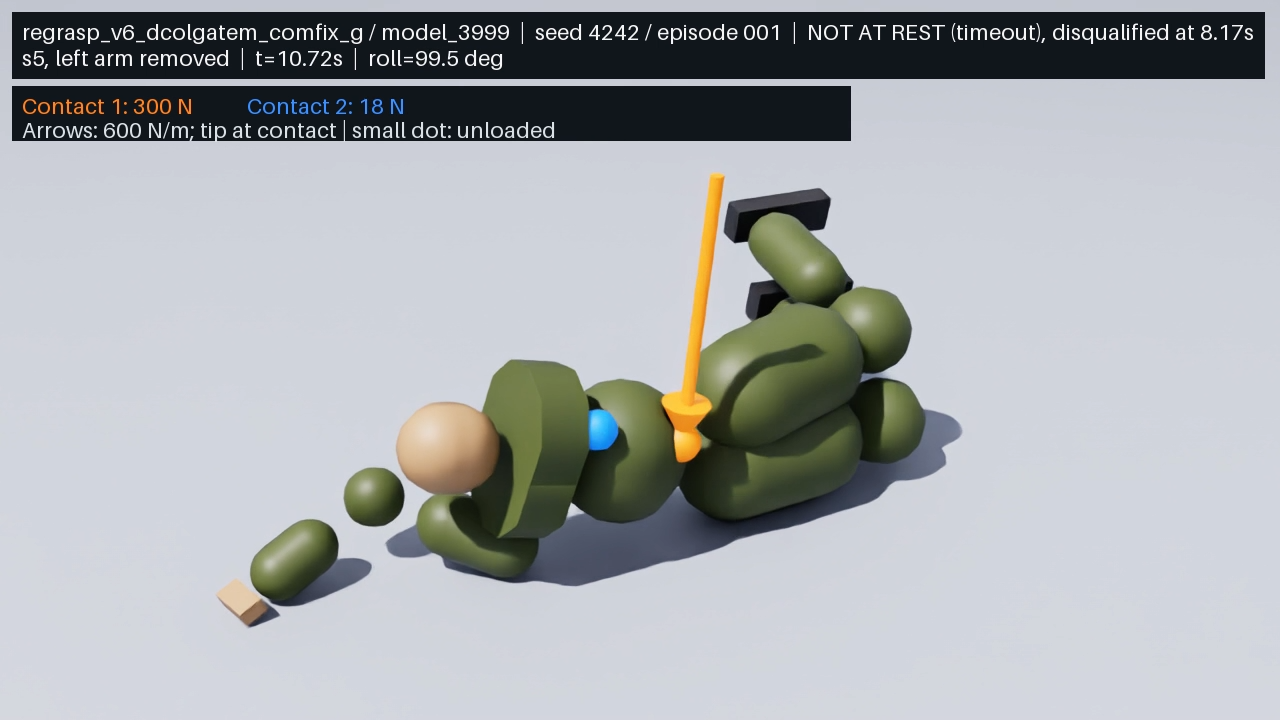} &
        \includegraphics[trim= 2cm 0cm 4cm 5.2cm, clip, width=0.18\textwidth]{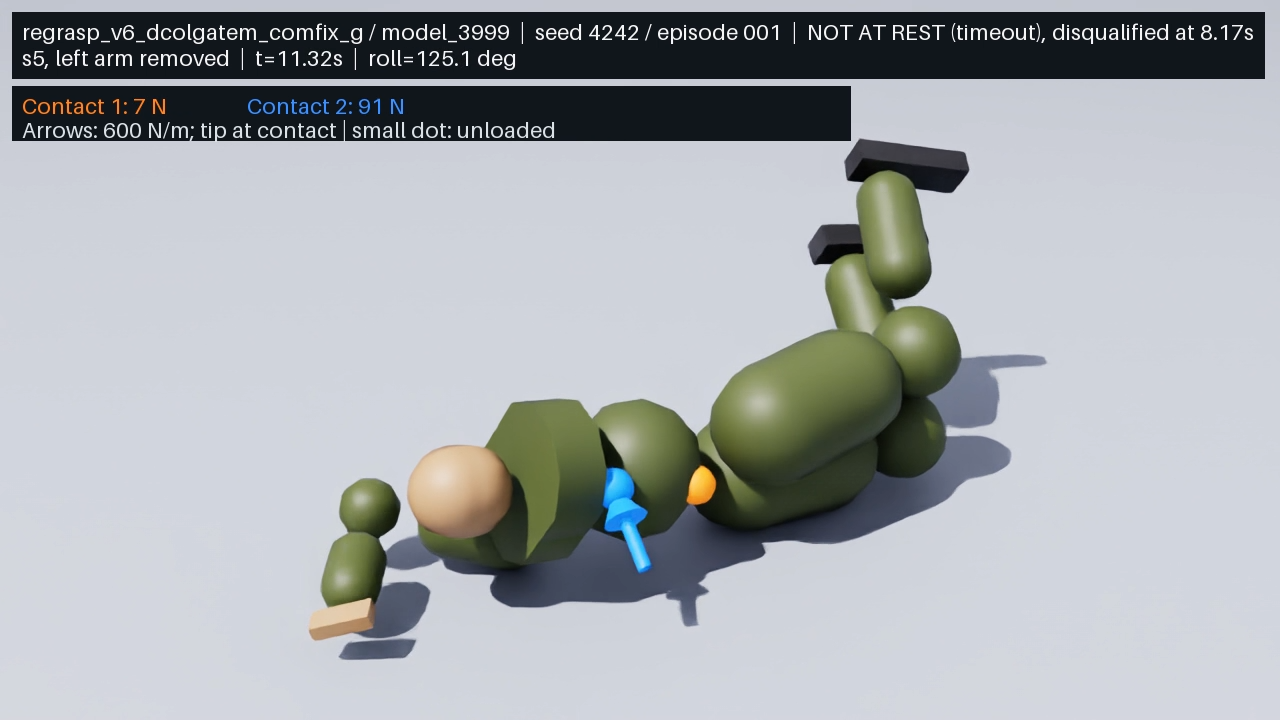} &
        \includegraphics[trim= 2cm 0cm 4cm 5.2cm, clip, width=0.18\textwidth]{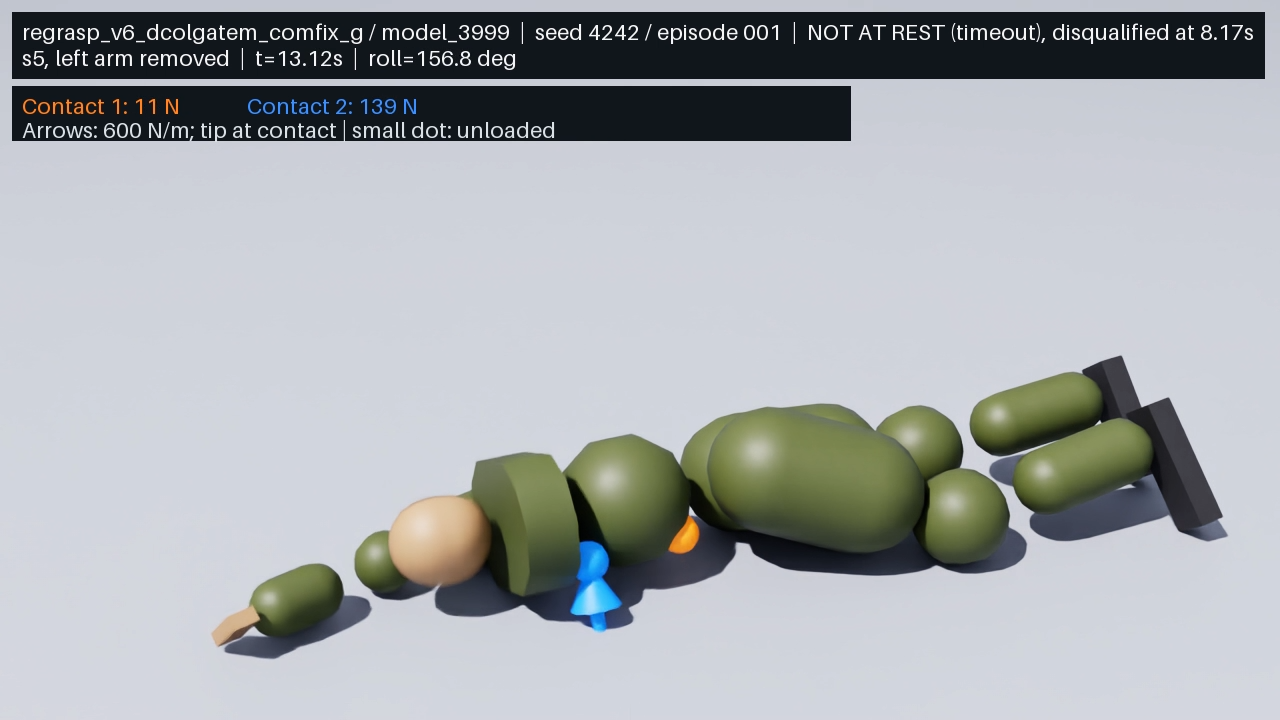} \\

        \rotatebox{90}{\textbf{Leg-R-1}} &
        \includegraphics[trim= 2cm 0cm 4cm 5cm, clip, width=0.18\textwidth]{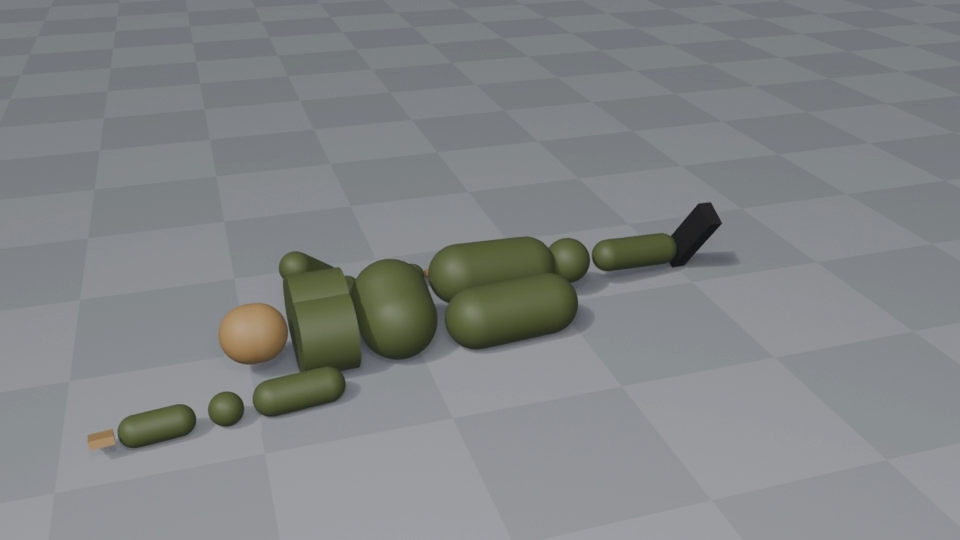} &
        \includegraphics[trim= 2cm 0cm 4cm 5cm, clip, width=0.18\textwidth]{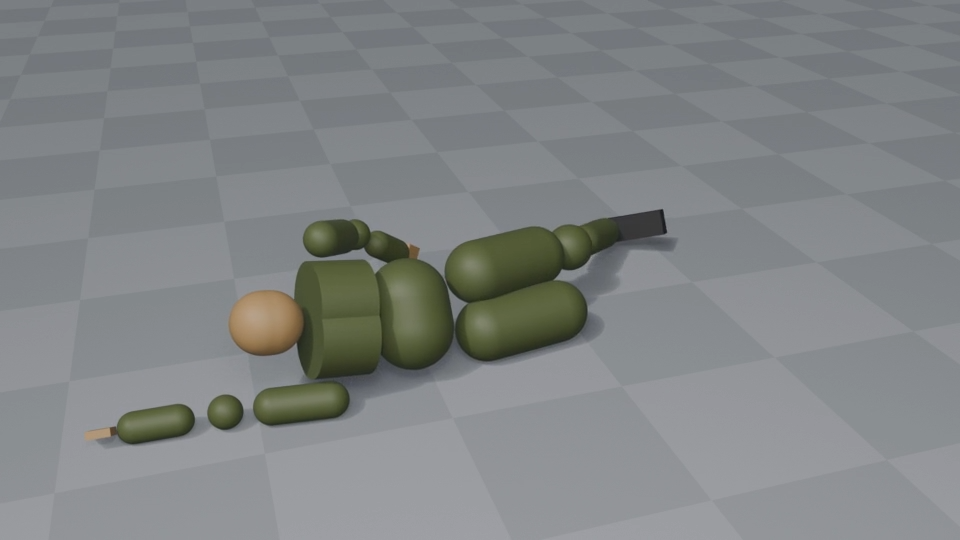} &
        \includegraphics[trim= 2cm 0cm 4cm 5cm, clip, width=0.18\textwidth]{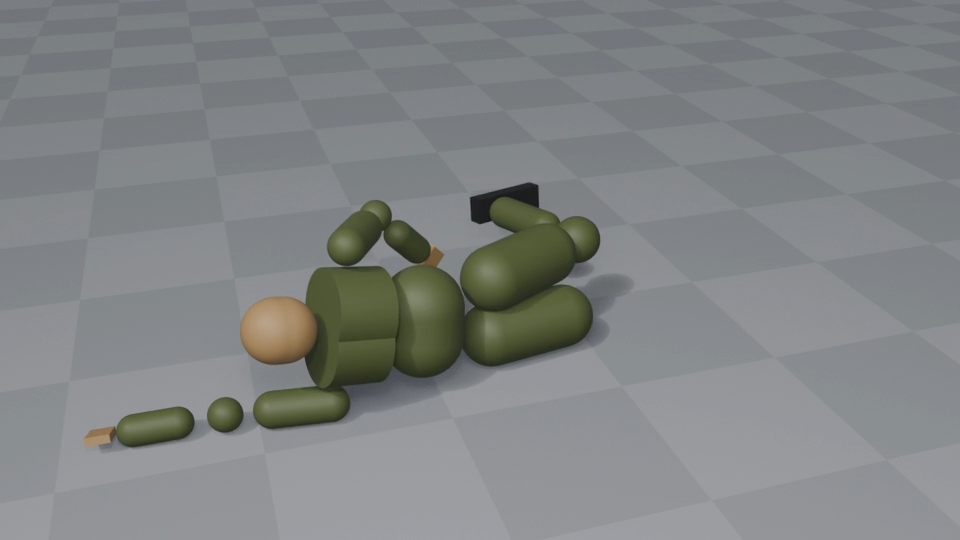} &
        \includegraphics[trim= 2cm 0cm 4cm 5cm, clip, width=0.18\textwidth]{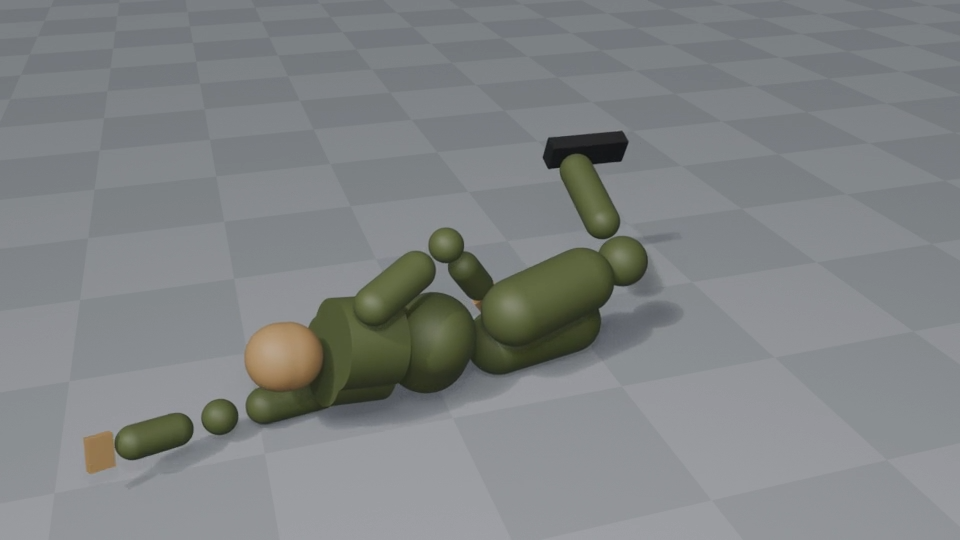} &
        \includegraphics[trim= 2cm 0cm 4cm 5cm, clip, width=0.18\textwidth]{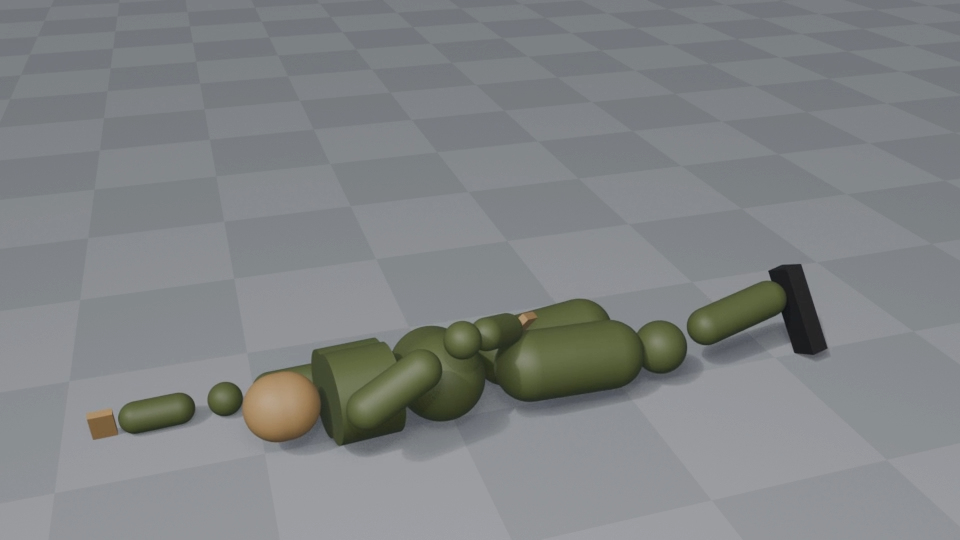} \\
    \end{tabular}
    \vspace{-1mm}
    \caption{Subject 5 was chosen for the baseline time series to show the visual differences in the policies across (A) Our Method with all safety considerations included, (B) Uniform choice of contact point, (C) Fixed Forces at the 95th percentile. The injury time series are shown in (D) left arm second joint amputation successful roll (E) right leg first joint amputation successful roll}
    \vspace{-6.5mm}
    \label{fig:comparison series}
\end{figure*}

\begin{table*}[t]
\vspace{3mm}
\centering
\caption{Baseline Comparisons}
\resizebox{\textwidth}{!}{%
\begin{tabular}{l c c c c c c c c c c}
\toprule
\textbf{Method} & \textbf{Reward}$\uparrow$ & \textbf{Success}$\uparrow$ & \textbf{Completion}$\%$ & \textbf{Rot. Speed (rad/s)}$\downarrow$ & \textbf{Body Align}$(^\circ)$$\downarrow$ & \textbf{Lift}$\uparrow$ & \textbf{Spin}$\uparrow$ & \textbf{Axial Rot.}$(^\circ)$ $\downarrow$ & \textbf{Force (N)}$\downarrow$ & \textbf{Force/Kg}$\downarrow$ \\
\midrule
\textbf{Ours} & $\mathbf{264.8}$ & $\mathbf{70.2}$ & \textbf{85.3} & $0.92$ & $7.63/15.10$ & $0.000$ & $-0.0022$ & $3.41/11.7$ & $227.1$ & $3.19$ \\
No DR & 242.5 & 63.7 & 81.3 & 1.29 & 7.83/16.74 & 0.0000 & $-0.0024$ & 2.33/9.4 & 231.7 & 3.22 \\
Uniform contacts & $-110.1$ & $4.00$ & $51.5$ & $5.00$ & $9.75/23.89$ & $-0.0022$ & $-0.0042$ & $5.55/18.6$ & $374.4$ & 5.25 \\
Fixed Force (5th) & $-15.9$ & $0.00$ & $0.00$ & $\mathbf{0.00}$ & $\mathbf{0.00/0.00}$ & $0.0000$ & $-0.0004$ & $\mathbf{0.01/0.01}$ & $\mathbf{16.9}$ & $\mathbf{0.24}$ \\
Fixed Force (50th) & $-12.5$ & $0.00$ & $3.00$ & $0.90$ & $2.42/4.10$ & 0.0000 & $\mathbf{-0.0002}$ & $0.84/2.10$ & 172.2 & 2.42 \\
Fixed Force (95th) & $-197.2$ & $0.00$ & $79.7$ & $7.04$ & $7.79 / 27.95$ & 0.0000 & $-0.0019$ & $4.39 / 27.4$ & $359.8$ & $3.82$ \\
\bottomrule
\end{tabular}
}
Comparison of the full learned policy against the fixed-force and uniform-contact baselines. \textbf{Bold} marks the best value per column.
\label{tab: trail_comp}
\vspace{-4mm}
\end{table*}

\begin{figure*}[t]
\vspace{2mm}
    \centering
        \includegraphics[width=\linewidth]{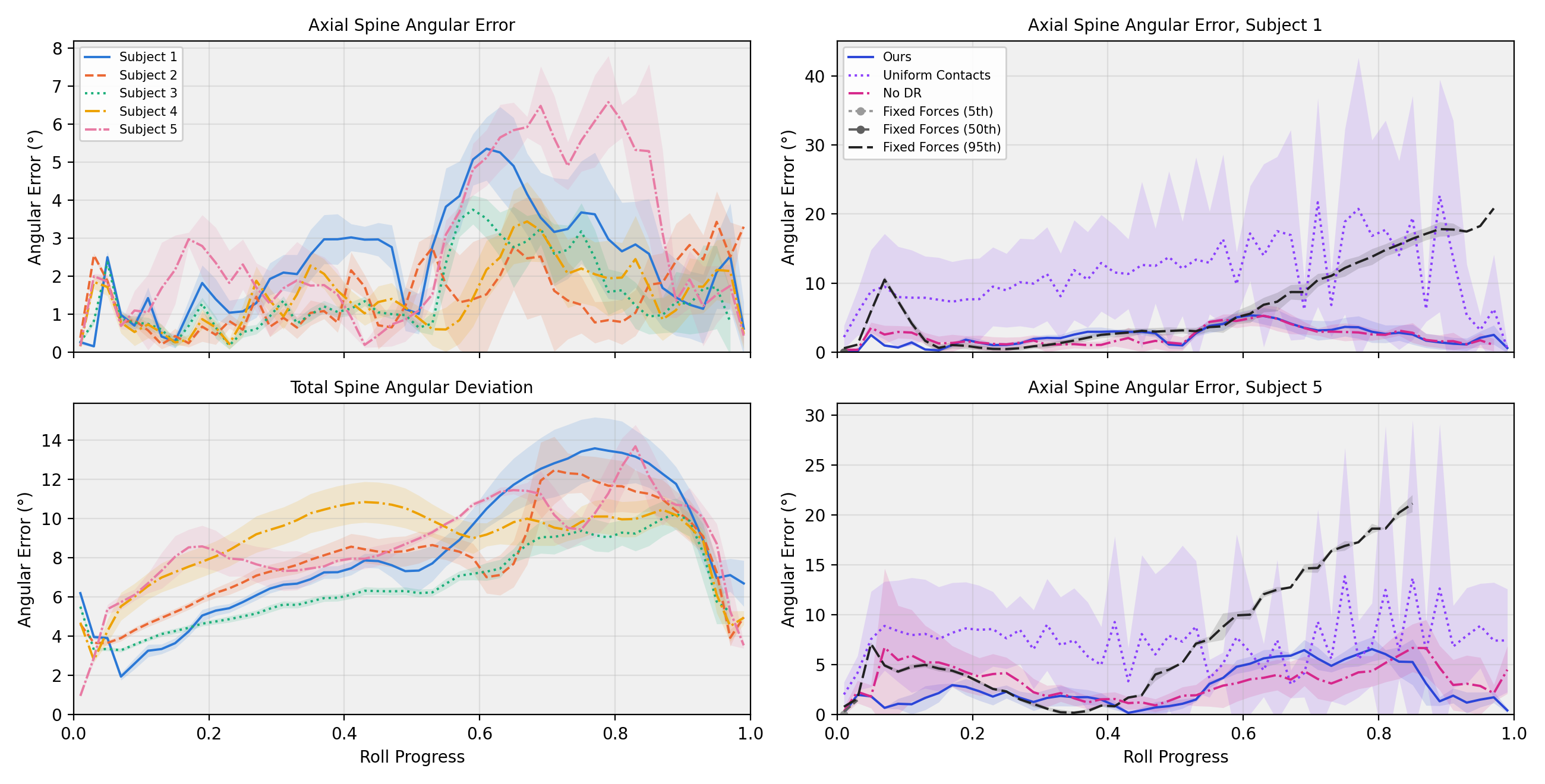}
        \label{fig:ours}
        \vspace{-4mm}
    \caption{Column 1: The total spinal deviation and axial spinal motion are shown based on all successful rolls from each subject. The most challenging subject 5 shows the highest average error in axial rotation; Column 2: spinal deviation successful (or all) rolls for all the baselines on subjects 1 and 5.}
    \label{fig: graphs}
    \vspace{-4mm}
\end{figure*}

\subsection{Meeting Safety Metrics}

The results of the training show a task completion rate of 85.3\% and a success rate, task completion, and acceptable safety metrics of 70.2 \% across all 5 models (Table~\ref{tab: trail_comp}). Our policy reaches the supine goal with a smooth force profile while holding axial twist below the $10\degree$ limit (Fig.~\ref{fig: graphs}(a)) and an overall deviation as defined in Sec \ref{Sec: alignment} of $16.79\degree$. Among the failures that still completed the task, peak axial twist reached $11.7\degree$ and peak spinal deviation reached $15.10\degree$. Meaning, even in failure cases, the axial alignment, which was the failure mode, stayed within $1.7\degree$ of the safety limit.

The most challenging deployment was subject 5 which features a mass distribution farthest outside the expected distribution in training. The model features a heavier lower body relative to it's upper body than other models which challenges the policy to meet the safety constraints. The policy's success means it generalized to individuals with weight distributions more similar to typical female proportions~\cite{schorr2018sex}. 

\subsection{Baseline Comparisons}
\label{sec:baselines}

 Initial tests to ensure a good policy included comparing the full policy training method against three controlled baselines that significantly simplify the solution to see the value of complexity and nuance in policy training. The baseline descriptions are listed below: 
 
\subsubsection*{\textbf{Fixed Forces (FF)}} The two contact bodies and force locations are fixed to clinically recommended placement, and constant force commands are applied (no learned modulation)---an open-loop push; Three tiers of magnitude [high, medium, low] are derived from the average force Gaussian distribution of successful rolls and taken from the 5th, 50th, and 95th percentile force magnitudes.

\subsubsection*{\textbf{Uniform Contact}} contact bodies are sampled uniformly at random from the candidate set each episode instead of higher preference given to clinical placement; 

\subsubsection*{\textbf{No Domain Randomization}} the policy is trained on a single nominal body only (subject 1), with proportions described in Fig.~\ref{fig:human_models}.

\subsubsection{Policy Comparison Across Baseline Models}

The results of the training performed on the 5 subjects are shown in Table~\ref{tab: trail_comp}. Each policy was deployed for 100 runs per subject, and the table values reflect averages across runs; some include average and peak values. For each condition, time-series diagnostics of the axial angle and deviation angle versus time (Fig.~\ref{fig: graphs}(b)), together with completion rate, peak spinal torsion and deviation, and peak force, are summarized in Table~\ref{tab: trail_comp}. The learned policy reaches the supine goal with a smooth force profile while holding torsion below the $10\degree$ limit (Fig.~\ref{fig: graphs}(a)). The \emph{fixed-force} baseline, lacking closed-loop modulation, either stalls before completing the roll or overshoots with large impulsive forces that drive torsion past the safe limit which is seen in Fig.~\ref {fig:comparison series} for the FF 95th percentile forces. The \emph{uniform-contact} baseline succeeds only intermittently, since randomly selected sites frequently fall outside the shoulder-girdle/pelvis pair required for a coordinated log roll, producing the shoulder-leads-pelvis torsion that the disagreement penalty is designed to suppress. It is also the only policy that produces measurable lift. The \emph{no-DR} baseline matches the full policy on the nominal body but degrades by producing higher peak forces across the anthropometric range and cannot reach the restrictive final pose for subject 5, confirming that randomization makes the strategy body-agnostic (generalizable). Fig.~\ref{fig:comparison series} shows the policies deployed on subject 5, the most out-of-distribution model, but does not show the trivial solutions that result in no movement from fixed forces at the 50th and 5th percentiles and no domain randomization.

\begin{table*}[t]
\centering
\caption{Ablation Study}
\label{tab:ablation}
\resizebox{\textwidth}{!}{%
\begin{tabular}{l c c c c c c c c c}
\toprule
\textbf{Configuration} & \textbf{Reward}$\uparrow$ & \textbf{Success}$\uparrow$ & \textbf{Completion}$\uparrow$ & \textbf{Body Align}$(^\circ)$$\downarrow$ & \textbf{Lift}$\uparrow$ & \textbf{Spin}$\downarrow$ & \textbf{Axial Rot.}$(^\circ)$$\downarrow$ & \textbf{Force (N)}$\downarrow$ & \textbf{Force/Kg}$\downarrow$ \\
\midrule
\textbf{Ours} & $\mathbf{264.8}$ & $\mathbf{70.2}$ & \textbf{85.3} & $7.63/15.10$ & $0.000$ & $-0.0022$ & $3.41/11.7$ & $227.1$ & $3.19$ \\
\midrule
w/o Gentle Set Down & $-$29.7 & 0.0 & 2.8 & 8.02/12.71 & 0.0000 & $-$0.0023 & 0.54/4.5 & 131.2 & 1.66 \\
w/o Spinal Align & $-$75.1 & 0.0 & 59.1 & 15.41/25.63 & 0.0000 & $-$0.0113 & 12.10/25.0 & 175.7 & 2.17 \\
w/o Spin Axis Alignment  & 276.3 & 65.0 & 69.7 & 8.76 / 16.66 & 0.0000 & $-$0.0036 & 3.32 / 9.9	 & 246.9 & 3.60\\
w/o Success Bonus & $-$56.1 & 0.0 & 35.6 & 10.17/27.24 & 0.0000 & $-$0.0058 & 2.95/18.6 & 164.1 & 2.00 \\
\bottomrule
\end{tabular}%
}
Each row removes one reward term from the full objective and retrains the domain-randomized task. \textbf{Bold} marks the best value per column.
\label{tab:ablation}
\vspace{-5mm}
\end{table*}

\begin{table*}[t]
\vspace{3mm}
\centering
\caption{Injury and Amputation Study}
\footnotesize
\setlength{\tabcolsep}{2pt}
\begin{tabular}{l *{8}{cc}}
\toprule
& \multicolumn{2}{c}{\textbf{Arm-L-1}} & \multicolumn{2}{c}{\textbf{Arm-R-1}} & \multicolumn{2}{c}{\textbf{Leg-L-1}} & \multicolumn{2}{c}{\textbf{Leg-R-1}} & \multicolumn{2}{c}{\textbf{Arm-L-2}} & \multicolumn{2}{c}{\textbf{Arm-R-2}} & \multicolumn{2}{c}{\textbf{Leg-L-2}} & \multicolumn{2}{c}{\textbf{Leg-R-2}} \\
\cmidrule(lr){2-3} \cmidrule(lr){4-5} \cmidrule(lr){6-7} \cmidrule(lr){8-9} \cmidrule(lr){10-11} \cmidrule(lr){12-13} \cmidrule(lr){14-15} \cmidrule(lr){16-17}
\textbf{Method} & Comp & Inj & Comp & Inj & Comp & Inj & Comp & Inj & Comp & Inj & Comp & Inj & Comp & Inj & Comp & Inj \\
\midrule
Ours (on Amp)        & 59 & - & 61 & - & 19 & - & 34 & - & 88 & - & 20 & - & 45 & - & 81 & - \\
Ours + Inj.  & 90 & $-4$ & - & - & 87 & $-3$ & - & - & 91 & 0 & - & - & 87 & 7 & - & - \\
Ours + Amp.  & 90 & 22 & 94 & 31 & 97 & 16 & 76 & 6 & 95 & - & 66 & - & 88 & - & 80 & - \\
\bottomrule
\end{tabular}\\
For all injury permutations: [Arm, Leg] followed by [L, R] for left or right, and [1, 2] designates which joint is injured or is the root of the amputation. The first column under each designated injury, ``comp,'' denotes the completed rolls, and ``inj'' denotes the \% reduction in injury motion.
\label{tab:inj_amp}
\vspace{-6mm}
\end{table*}

\subsection{Reward Ablation Study}
\label{sec:ablation}
Every reward component can be disabled independently. We retrain the domain-randomized task with each component removed one by one; Table~\ref{tab:ablation} reports policy deployment on the 5 models, with each model run 100 times. This answers, per term, whether it is (a) necessary for task completion, (b) necessary for safety, or (c) redundant given the other terms. Notably, some terms that raise the raw reward  (e.g.\ removing the force penalty) do so by spending far more force for the full reward, confirming that the term trades nominal reward for the gentler, safer strategy the task actually requires. Similarly, omitting the smoothness penalty yields comparable safety numbers; however, it reduces completion and supports instantaneous changes in force, which are physically damaging to a patient. 
\vspace{-.01mm}
To evaluate robustness to real field casualties, we test the policy on bodies that depart from the intact nominal model in two ways: (1) prescribing a joint that is indicated as injured; (2) removing full and partial limbs. Both scenarios include an additional reward that penalizes angular-position changes of the injured joint or the amputation root joint throughout the roll, so a single conditioned policy learns to route forces to reduce turbulence at the injury site throughout the roll.
\vspace{-.5mm}
\subsection{Minimize injured Joint Motion}
\vspace{-.5mm}
In the \emph{injury-conditioned} condition, an injured joint is sampled each episode from [hip, shoulder, knee, elbow], encoded as a one-hot input. The injuries are applied to the side of the body not being rolled against the ground (the left side). For each condition, we report the completion rate and compare the injury-conditioned policy with the unconditioned policy to quantify how much conditioning reduces motion at the protected joint. Table~\ref{tab:inj_amp} shows the results with injury conditions. The policy is able to maintain a high completion rate but does not make any meaningful reduction of limb motion due to the uncontrollability of the limbs during the roll. The only reported improvement was a 7\% reduction for the left leg.

\subsubsection{Amputations}

In the \emph{amputation} condition, one limb segment is removed from the model (single amputation, with the amputation site provided to the policy), reflecting that amputations are among the most common blast injuries; the roll must still reach supine without the missing limb as a usable handle or counterweight. In Table~\ref{tab:inj_amp}, compared to the baseline method not trained for amputations, the roll completion rate increases significantly. The largest increase was 78\% more completed rolls for the left lower-leg amputation. Because removing partial limbs decreased uncertainty about limb position, this policy was able to reduce shoulder and hip joint motions significantly, up to 31\%.

\section{Discussion and Conclusion}
\vspace{-1.5mm}
In this work, a biomechanically accurate human model is used to define a policy for rolling, informing future designs for robots capable of high force output and delicate manipulation. While the model and simulation provide meaningful results and interesting insights, this work is limited in its real-world transition. In the pHRI space, real-world experiments often come at a cost and with limitations. Deploying these algorithms with human users would first require formal safety guarantees. To work with real-world, human-like objects, we sacrifice accuracy and representative diversity. Most human-shaped phantoms lack the attributes needed to serve as a platform for transferable policy results. Further, designing a single representative phantom would not capture the natural, impactful variation in human features such as weight distribution, height, and anatomical structure.
While real-world evaluation of our policy is a critical future direction, this work provides insights that can guide necessary developments.
\vspace{-2mm}
\section{Acknowledgments}
Claude Code was utilized to help implement the described code/algorithms and scripts to graph the visualized data. Claude was used to provide feedback on the continuity and clarity of the writing, but it was not used to generate any of the prose.

\balance
\bibliographystyle{ieeetr}
\bibliography{refs}
\balance


\end{document}